\pdfoutput=1

\documentclass[11pt]{article}

\usepackage{acl}

\usepackage{times}
\usepackage{latexsym}
\usepackage{CJKutf8}

\usepackage[T1]{fontenc}

\usepackage[utf8]{inputenc}

\usepackage{microtype}
\usepackage{tabularx}
\usepackage{amsmath}
\usepackage{amssymb}
\usepackage{dsfont}
\usepackage{xcolor}
\usepackage{arydshln}
\usepackage{tikz}
\usetikzlibrary{arrows, decorations.text, shapes.geometric, positioning, decorations.pathreplacing, calligraphy}
\usepackage{pgfplots}
\usepackage{pgfmath}
\usepackage{pgffor}
\pgfplotsset{compat=1.17}
\usepackage{subcaption}
\usepackage{mathtools}
\usepackage{multirow}
\usepackage{booktabs}
\usepackage{pifont}
\usepackage{ifthen}
\usepackage{enumitem}
\definecolor{Fix}{RGB}{0,0,0}
\definecolor{Fix2}{RGB}{0,0,0}

\title{Towards Fine-Grained Information:\\Identifying the Type and Location of Translation Errors}

\author{
    Keqin Bao$^{1,3}$\thanks{~~Equal contribution. Work was done when Keqin Bao and Yu Wan were interning at DAMO Academy, Alibaba Group.}~~~Yu Wan$^{2,3*}$~~~Dayiheng Liu$^3$~~~Baosong Yang$^3$~~~\textbf{Wenqiang Lei}$^4$\\\textbf{Xiangnan He}$^1$~~~\textbf{Derek F. Wong}$^2$~~~\textbf{Jun Xie}$^3$\\
    \vspace{0.25ex}
    $^1$University of Science and Technology of China~~~~~~$^2$NLP$^2$CT Lab, University of Macau\\
    $^3$DAMO Academy, Alibaba Group~~~~~~$^4$National University of Singapore\\
    \vspace{-0.75ex}
    \small{\tt{baokq@mail.ustc.edu.cn~~~nlp2ct.ywan@gmail.com}}\\
    \vspace{-0.75ex}
    \small{\tt{\{liudayiheng.ldyh,yangbaosong.ybs,qingjing.xj\}@alibaba-inc.com}}\\
    \vspace{-0.75ex}
    \small{\tt{wenqianglei@gmail.com}~~~\tt{xiangnanhe@gmail.com}~~~\tt{derekfw@um.edu.mo}}
}

\begin{document}

\maketitle



\begin{abstract}

Fine-grained information on translation errors is helpful for the translation evaluation community.
Existing approaches can not synchronously consider error position and type, failing to integrate the error information of both.
In this paper, we propose Fine-Grained Translation Error Detection (FG-TED) task, aiming at identifying both the position and the type of translation errors on given source-hypothesis sentence pairs.
Besides, we build an FG-TED model to predict the \textbf{addition} and \textbf{omission} errors -- two typical translation accuracy errors.
First, we use a word-level classification paradigm to form our model and use the shortcut learning reduction to relieve the influence of monolingual features.
Besides, we construct synthetic datasets for model training, and relieve the disagreement of data labeling in authoritative datasets, making the experimental benchmark concordant.
Experiments show that our model can identify both error type and position concurrently, and gives state-of-the-art results on the restored dataset.
Our model also delivers more reliable predictions on low-resource and transfer scenarios than existing baselines.
The related datasets and the source code will be released in the future.

\end{abstract}

\section{Introduction}
\label{introduction}

Fine-grained information of translation errors is important for the translation evaluation community~\cite{freitag2021experts,vamvas-sennrich-2022-little}. 
Given the sentence pair including translated hypothesis (\textsc{Hyp}) and source input (\textsc{Src}), models are required to provide fine-grained error information, such as the error type and the error position (\textit{e.g.}, addition error ``\textcolor{red}{last month}'' in Table~\ref{table_annotaion}).
Approaches like word-level quality estimation~\citep[word-level QE,][]{kim-etal-2017-predictor,basu-etal-2018-keep} and critical error detection~\citep[CED,][]{specia-etal-2021-findings} models can predict detailed information on translation errors.
Compared to sentence-level QE models~\cite{ranasinghe2021} which give overall translation quality scores, the outputs of word-level QE and CED models can better help researchers know where they are located and why the translated words are wrong. 
Besides, the detailed illustrations can also help analyze the translation models and judge the translation quality, offering explainability for the related studies on machine translation~\cite{kim-etal-2017-predictor,specia-etal-2018-findings}.

\begin{CJK*}{UTF8}{gbsn}
\begin{CJK}{UTF8}{gbsn}
    \begin{table}[t]
        \centering
        \small
        \scalebox{0.65}{
            \begin{tabular}{lcc}
                \toprule
                 \textbf{\textsc{Dataset}} & \textbf{\textsc{Src}} & \textbf{\textsc{Hyp}} \\
                \midrule
                \multicolumn{3}{c}{\textit{Golden Translation}} \\
                \cdashline{1-3}[1pt/2.5pt]\noalign{\vskip 0.5ex}
                & 学校和幼儿园开学了。 & Schools and kindergartens opened. \\
                \midrule
                
                \multicolumn{3}{c}{\textit{Erroneous Translation}} \\
                \cdashline{1-3}[1pt/2.5pt]\noalign{\vskip 0.5ex}
                & 学校\underline{和幼儿园}开学了。 & Schools opened \underline{last month}. \\
                \midrule
                \multicolumn{3}{c}{\textit{Dataset Labeling Format}} \\
                \cdashline{1-3}[1pt/2.5pt]\noalign{\vskip 0.5ex}
                \textbf{MQM'20} & 学校和幼儿园开学了。 & Schools \textcolor{cyan}{\underline{and kindergardens}}  opened \textcolor{red}{\underline{last month}}. \\
                \textbf{MQM'21} & 学校\textcolor{cyan}{\underline{和幼儿园}}开学了。 & Schools  opened  \textcolor{red}{\underline{last month}}.\\
                
                \bottomrule
                 
            \end{tabular}
        }
        \vspace{-0.5em}
        \caption{A toy illustration of the \textcolor{red}{addition} and \textcolor{cyan}{omission} error and the label format of available datasets. Some omission errors from MQM'20 dataset~\cite{freitag2021experts} are labeled at the target side\textcolor{Fix2}{, where t}he missing semantics is complemented with additional tokens (``\textcolor{cyan}{and kindergardens}''). We \textcolor{Fix2}{follow the setting of MQM'21 dataset and} relabel the location of Chinese-English omission errors to remove such \textcolor{Fix2}{format disagreement}. See \textcolor{Fix2}{Appendix} \S\ref{appendix.human} for more details.}
        \label{table_annotaion}
        \vspace{-1em}
    \end{table}
\end{CJK}
Despite the success of word-level QE~\cite{specia-etal-2020-findings-wmt} and CED approaches~\cite{freitag-etal-2021-results}, they can not concurrently identify the types and locations of wrongly translated spans, failing to integrate the information of both.
To bridge this gap, in this paper, we propose fine-grained translation error detection (FG-TED) task: the model should not only identify the wrongly translated spans, but also predict what error type each span belongs to.
In practice, we prioritize the detection of \textbf{addition} and \textbf{omission} translation errors, two typical types of translation errors in real-world applications.\footnote{For other translation error types, we discuss them in \S\ref{conclusion}.}
As the ``Erroneous'' example in Table~\ref{table_annotaion}, the span of \textsc{Hyp} is labeled as \textcolor{red}{addition error} if its semantics is excluded in the \textsc{Src} (\textit{i.e.}, ``\textcolor{red}{last month}''), and the \textcolor{cyan}{omission error} means that the semantics of span in \textsc{Src} is omitted (\textit{i.e.}, ``\textcolor{cyan}{和幼儿园}'').
\end{CJK*}

To build an FG-TED model, we mainly face challenges from two aspects, \textit{i.e.}, model and data.
For the former, recent QE methods involve pre-trained language models (PLMs) to extract the semantic representations of \textsc{Hyp} and \textsc{Src} sentence~\cite{kepler-etal-2019-openkiwi,ranasinghe2021}.
However, most PLMs~\citep[\textit{e.g.}, \textsc{XLM-R},][]{conneau2019unsupervised} are trained by Masked Language Modeling (MLM) objective, focusing on modeling the sentence fluency and grammatical correctness~\cite{behnke-etal-2022-bias}.
We find that, those PLMs tend to employ ``shortcuts'' during fine-tuning -- they mainly use monolingual features during learning, failing to utilize cross-lingual ones when collecting fine-grained translation error information.
To alleviate this, we consider two solutions as follows:
1) As to the PLM backbone, we utilize those ones which are enhanced with cross-lingual semantic alignments.
2) As to the model training, we introduce shortcut learning reduction (SLR) loss to prevent the model from overly using monolingual information during training.
Experimental results also verify that, combining both the strategies above can further improve the FG-TED model performance.

For the latter, we mainly face two difficulties.
On one hand, to the best of our knowledge, the Multidimensional Quality Metrics (MQM) dataset~\cite{freitag2021experts} is the authoritative dataset annotating both the types and the locations of translation errors.
Yet, as in Table~\ref{table_annotaion}, the omission errors of some examples are denoted on \textsc{Hyp}, which shows disagreement of MQM benchmarks across multiple years~\cite{vamvas-sennrich-2022-little}.
On the other hand, the number of samples containing addition/omission errors in the MQM datasets is insufficient.\footnote{The statistics can be seen in Table~\ref{table_statistic}.}
Therefore, we consider collecting synthetic examples involving addition/omission errors for model training. 
Experiments show that, after relabeling the omission errors in MQM'20 Chinese-English (Zh-En) datasets for reliable evaluation,\footnote{Due to the limitation of linguistic expertise of all authors, we only relabeled Zh-En examples.}
\textcolor{Fix2}{our model performs better than existing baselines, and it can achieve better results on the sub-tasks of FG-TED, \textit{i.e.}, word-level QE and CED.}

\section{Related Work}
\begin{figure}[t]
    \centering
    \scalebox{0.8}{
    \includegraphics[width=\columnwidth]{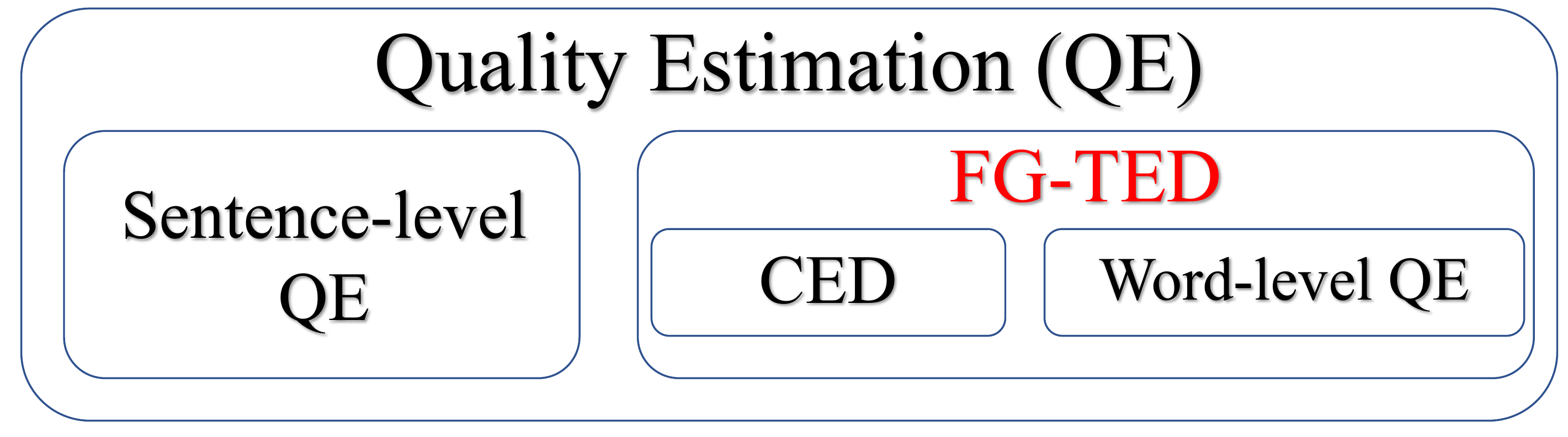}
    }
    \vspace{-1em}
    \caption{Illustration on fine-grained translation error detection (FG-TED) task and existing quality estimation (QE) tasks. Critical Error Detection (CED) and word-level QE can be regarded as FG-TED subtasks.}
    \label{figure.example}
    \vspace{-1em}
\end{figure}

\paragraph{Related Tasks}
Word-level QE~\cite{specia-etal-2018-findings} is the most similar task to FG-TED.
Most existing approaches like \textsc{IST-Unbabel}~\cite{zerva-etal-2021-ist} and \textsc{TransQuest}~\cite{ranasinghe2021} are supervised -- models are trained on the datasets~\citep[\textit{e.g.}, MLPE-QE,][]{fomicheva-etal-2022-mlqe} where each token is labeled as either ``[OK]'' or ``[BAD]''~\cite{fonseca-etal-2019-findings}, and use ``[GAP]'' in the \textsc{Hyp} side to indicate whether there is omission error between adjacent words.
Nevertheless, the predictions lack more details of translation errors,~\textit{i.e.}, what error type each ``[BAD]'' token or space belongs to.
Besides, the CED task also makes a preliminary attempt to arrange fine-grained analyses on translation errors.
It defines some accuracy errors (\textit{e.g.}, numeric) as the critical translation errors, and requires the models to deliver binary predictions to identify whether critical errors are involved in the example or not~\cite{specia-etal-2021-findings,amrhein2022aces}. 
However, CED models can not exactly locate what words are wrongly translated.
In summary, as shown in Figure~\ref{figure.example}, both word-level QE and CED tasks can be regarded as subtasks of our FG-TED task.

\paragraph{Available Dataset}

\textcolor{Fix2}{
The FG-TED task requires the test set examples fulfilled with both the type and location of translation errors for reliable evaluation.
The MQM'20 and MQM'21 are two authoritative datasets provided by the organizers of WMT~\cite{freitag-etal-2021-results}.
They include such detailed translation error information, but primarily serve as a benchmark for sentence-level tasks, \textit{i.e.}, giving an overall score to describe the quality of \textsc{Hyp}~\cite{xu2022not, rei2022comet}.
Nevertheless, limited by the disagreement of labeling format (Table~\ref{table_annotaion}) and the amount of data examples (Table~\ref{table_statistic}), utilizing MQM datasets for evaluation is not reliable.
In practice, we relabel the Zh-En examples from the MQM'20 dataset to preserve the consistency between MQM'20 and MQM'21 Zh-En subset, facilitating a more reliable evaluation benchmark.
}

\paragraph{Related Methods} Several existing works have explored the topic of fine-grained translation error information.
\newcite{zhou2021detecting} constructed the pseudo examples containing translation hallucination errors, and fine-tuned a PLM to detect hallucinated translations.
However, their approach fails to identify translation errors at the source side, because the hallucination errors only exist in the \textsc{Hyp} sentence.
\newcite{vamvas-sennrich-2022-little} adapted the idea of contrastive conditioning~\cite{vamvas-sennrich-2021-contrastive} to predict error spans.
They use a two-stage approach -- adapting a dependency parser to collect text spans and calculating the decrease in translation probabilities after removing each span.
Nevertheless, when detecting errors, this approach requires a large amount of time when processing long sentences.
Compared to those approaches, our model can not only concurrently detect addition and omission errors, but also ease the application following an end-to-end paradigm.

\paragraph{Shortcut Learning Reduction}

Shortcut learning reduction (SLR) aims at preventing the model from bridging the wrong relationship between features and labels.
\newcite{mahabadi2020end} adapted debiased focal loss (DFL) into object detection and natural language inference (NLI) tasks to alleviate the data-level bias between features and labels~\cite{lin2017focal}. 
\newcite{ganin2015unsupervised} and \newcite{belinkov-etal-2019-adversarial} used a gradient reversal layer (GRL) to help models ignore those spurious correlations in data on domain adaption and NLI tasks.
For the QE tasks, \newcite{behnke-etal-2022-bias} attempt to adapt the above two methods into sentence-level QE for bias mitigation, while the performance improvement is not stable across different translation directions.
To our view, DFL introduced a biased model to learn the bias and reweight the loss of the base model.
However, the NLI task is different from QE tasks, where it is difficult to isolate the bias from \textsc{Hyp} sentence~\cite{behnke-etal-2022-bias}.\footnote{Refer to \S\ref{appendix.dfl} for more analyses.}
Besides, GRL is unstable due to the simultaneous optimization of adversarial tasks, which easily leads to model confusion~\cite{ganin2016domain}.
By contrast, our proposed SLR uses a single model and restricts simultaneous optimization, which can both enhance the usage of cross-lingual features and stabilize the model training.

\section{Methods}
\begin{figure}[t]
    \centering
    \includegraphics[width=1.0\columnwidth, height=0.5\columnwidth]{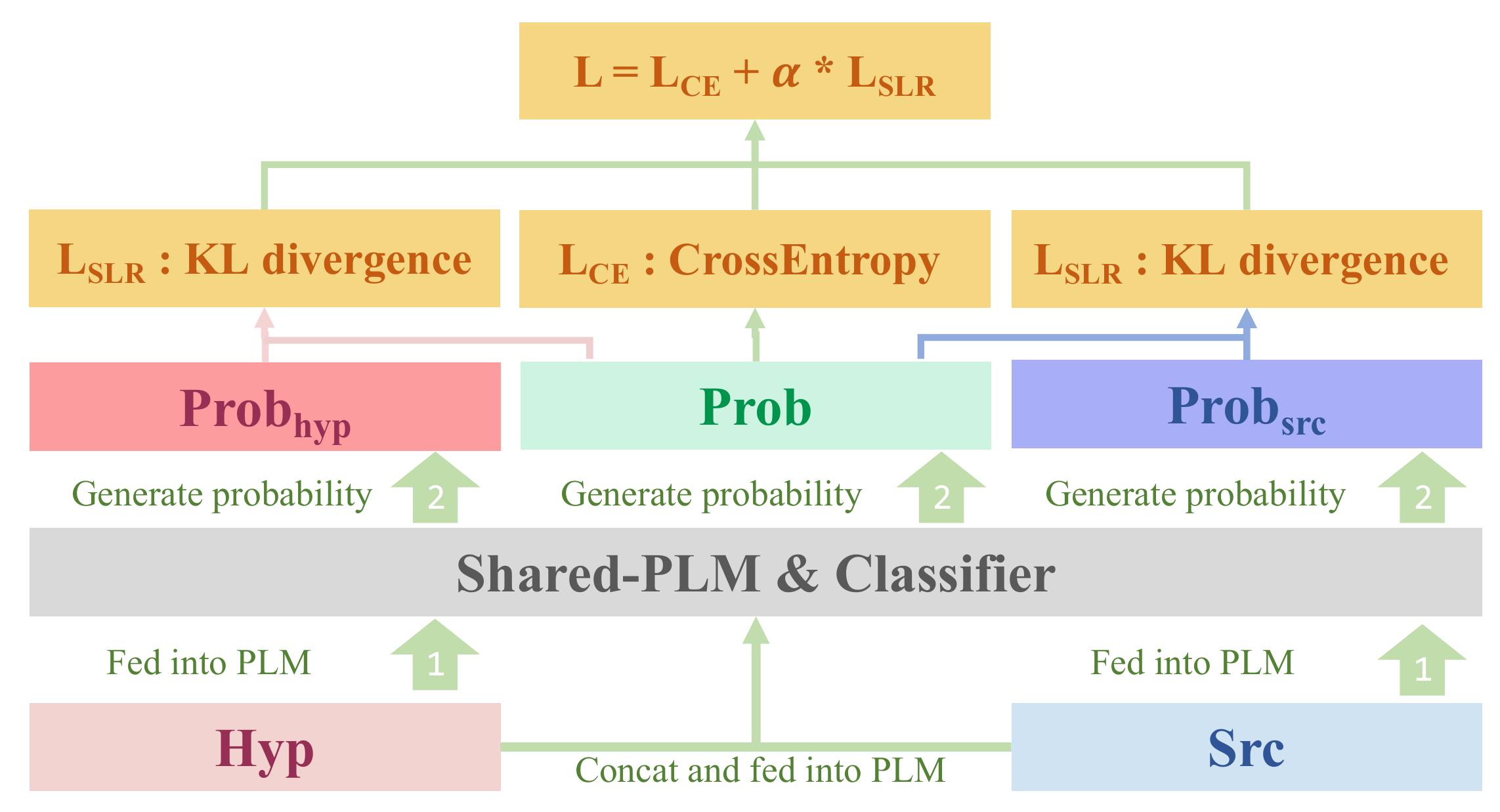}
    \caption{Illustration of our model architecture. Our model involves two shared modules: a PLM and a classifier. During training, our model first derives word-level classification results on the sentence-pair input, and reduces shortcut learning on monolingual inputs.
    }
    \label{fig.pipeline}
    \vspace{-1em}
\end{figure}
\label{section.method}
We first give the formulation of our FG-TED task (\S\ref{subsection.task}), then the proposed method 
(\S\ref{subsection.model}) and synthetic data construction (\S\ref{subsection.data}) are introduced.


\subsection{Task Formulation}
\label{subsection.task}
We define our task as a word-level classification problem. 
Specifically, given \textsc{Hyp} sentence $ \mathbf{h} = [h_1, h_2, \cdots, h_m] $ and \textsc{Src} sentence $ \mathbf{s} = [s_1, s_2, \cdots, s_n] $, models need to give predictions $ \mathbf{y} = [y_{h_1}, y_{h_2}, \cdots, y_{h_m}, y_{s_1}, y_{s_2}, \cdots, y_{s_n}] $ for all words.
In this research, we mainly consider addition and omission errors -- typical translation errors in practice~\cite{vamvas-sennrich-2022-little}.

\subsection{Model Architecture}
\label{subsection.model}

Our model architecture is shown in Figure~\ref{fig.pipeline}. 
The \textsc{Hyp} and \textsc{Src} are concatenated into a segment $\mathbf{x} = [\mathbf{h};~\mathbf{s}]$, which is fed into the PLM to obtain the embeddings of all tokens.
All embeddings are served as the input to a binary classifier.
The FG-TED model identifies the error words if the corresponding predictions are positive.
Otherwise, the words with negative scores are correctly translated.
The word-level classification learning objective of our model is to minimize the cross-entropy loss:
\begin{align}
\mathbf{P} & = \mathrm{softmax}(f(\mathbf{x}, \theta)) \in \mathbb{R}^{(m+n) \times 2}\\
\mathcal{L}_{CE} & = - \sum\limits_{i = 1}^{m + n}\sum\limits_{j = 0}^{1} \mathds{1}_{\mathbf{y}_i = j} \cdot log(\mathbf{P}_{i,j}),
\end{align}
where $\mathbf{y}$ is the ground truth, $f(\cdot, \cdot)$ represents the FG-TED model, and $\theta$ is the parameter set. 
Further, the erroneous words are labeled as omission and addition errors if they are on the source and  target sides, respectively.

\paragraph{PLM Selection}

We believe that, the cross-lingual alignments can vastly help our model, because they can deliver aligned semantics for \textsc{Hyp} and \textsc{Src} sentences.
To make full use of cross-lingual information for our FG-TED model, we apply~\textsc{InfoXLM}~\cite{chi-etal-2021-infoxlm}, which is enhanced with cross-lingual semantic information during the pre-training phase, as PLM backbone instead of conventional choice XLM-R~\cite{conneau2020unsupervised}.\footnote{We also testify the performances of different backbones pre-trained with other objectives in \S\ref{section.analysis}.}


\paragraph{Shortcut Learning Reduction}
\label{subsection.loss}

The existing PLMs usually contribute many monolingual features to fine-tuning, such as fluency and grammar~\cite{behnke-etal-2022-bias}.
Unavoidably, we find that, our FG-TED model also easily falls into this ``shortcut'': learning to predict the labels mainly based on monolingual information they learned during the pre-training period.\footnote{Empirical evidence can be found in \S\ref{section.exp}.}
To eliminate this effect, we guide the FG-TED model to distinguish the predictions on con catenated input ($\mathbf{x}$) and those on single-sentence inputs ($\mathbf{h}$ and $\mathbf{s}$).
In detail, apart from feeding $\mathbf{x}$ into our model, we additionally collect the predictions via taking $\mathbf{h}$ or $\mathbf{s}$ individually:
\begin{align}
    \mathbf{Q} & = \mathrm{softmax}([f(\mathbf{h, \theta}); f(\mathbf{s, \theta})]).
\end{align}
Then, we use KL-divergence loss to help push apart the distribution of $\mathbf{Q}$ away from that of $\mathbf{P}$.
The learning objective for SLR is formulated as:
\begin{align}
    \mathcal{L}_{SLR}(\theta) & = - \sum\limits_{i = 1}^{m + n}  \hat{\mathbf{Q}}_{i,\cdot} \cdot \mathrm{KL\text{-}div}(\mathbf{Q}_{i,\cdot} || \hat{\mathbf{P}}_{i,\cdot}).
\end{align}
\textcolor{Fix2}{As seen, two model input formats are leveraged in our SLR strategy, namely the concatenated input ($\mathbf{x}$) and single inputs ($\mathbf{h}$ and $\mathbf{s}$).
The objective of our SLR strategy is to maximize the difference between the corresponding representations $\mathbf{Q}$ and $\mathbf{P}$, preventing the model from solely relying on monolingual features during training.}

\textcolor{Fix2}{Note that, to preserve the efficiency of our approach, we share the PLM to obtain $\mathbf{Q}$ and $\mathbf{P}$.
However, we observe that back-propagating the gradients of $\mathbf{P}$ and $\mathbf{Q}$ simultaneously can lead to model collapse.
Our conjecture is that optimizing model parameters with KL-divergence loss delivers unstable learning.
To help the model stabilize the training, we first use detached sentence-pair predictions $\hat{\mathbf{P}}$ to obtain KL-divergence loss.}
Besides, for \textcolor{Fix2}{the wrongly translated tokens} whose probabilities in $\mathbf{Q}$ are strongly positive, \textcolor{Fix2}{our} model \textcolor{Fix2}{is} easily over-confident in \textcolor{Fix2}{its own} predictions~\cite{pereyra-etal-2017-regularizing,gao-etal-2020-towards}.
To alleviate \textcolor{Fix2}{such} over-fitting \textcolor{Fix2}{problem}, we apply the detached representations $\hat{\mathbf{Q}}$ \textcolor{Fix2}{as a multiplicator} for model regularization.

Finally, our loss function can be written as:
\begin{equation} \label{eq3}
\mathcal{L}=  \mathcal{L}_{CE} + \alpha \mathcal{L}_{SLR},
\end{equation}
where $\alpha$ is a hyper-parameter to balance between preserving the cross-lingual information and inhibiting the use of monolingual features.\footnote{\textcolor{Fix2}{In this research, we set $\alpha$ as 0.05 and 0.1 for \textsc{InfoXLM} and \textsc{XLM-R} backbone for all experiments, respectively. Analysis for tuning this hyper-parameter is in Appendix~\S\ref{appendix.parameter}.}}



\subsection{Data Collection}
\label{subsection.data}

Currently, the scarcity of labeled data hinders the research from error detection~\cite{vamvas-sennrich-2022-little}.
In this research, we alleviate this problem in two ways:
1) \textcolor{Fix2}{We construct synthetic data to help train the FG-TED model;}
2) We collect the MQM dataset~\cite{freitag2021experts} which includes English-German (En-De) and Zh-En examples.
Then, we relabel the Zh-En examples with well-formatted addition/omission error labels, making the dataset reliable for evaluation.

\paragraph{Synthetic Data Construction}
Inspired by~\newcite{sellam-etal-2020-bleurt,zhou2021detecting}, after collecting parallel corpora, we insert multiple mask tokens into one sentence, and fill them to construct an addition or omission error.
In detail, as shown in Figure~\ref{figure.data}, the pipeline for obtaining such data consists of the following steps:

\begin{figure}[t]
    \centering
    \scalebox{1.0}
    {
        \includegraphics[width=1\columnwidth]{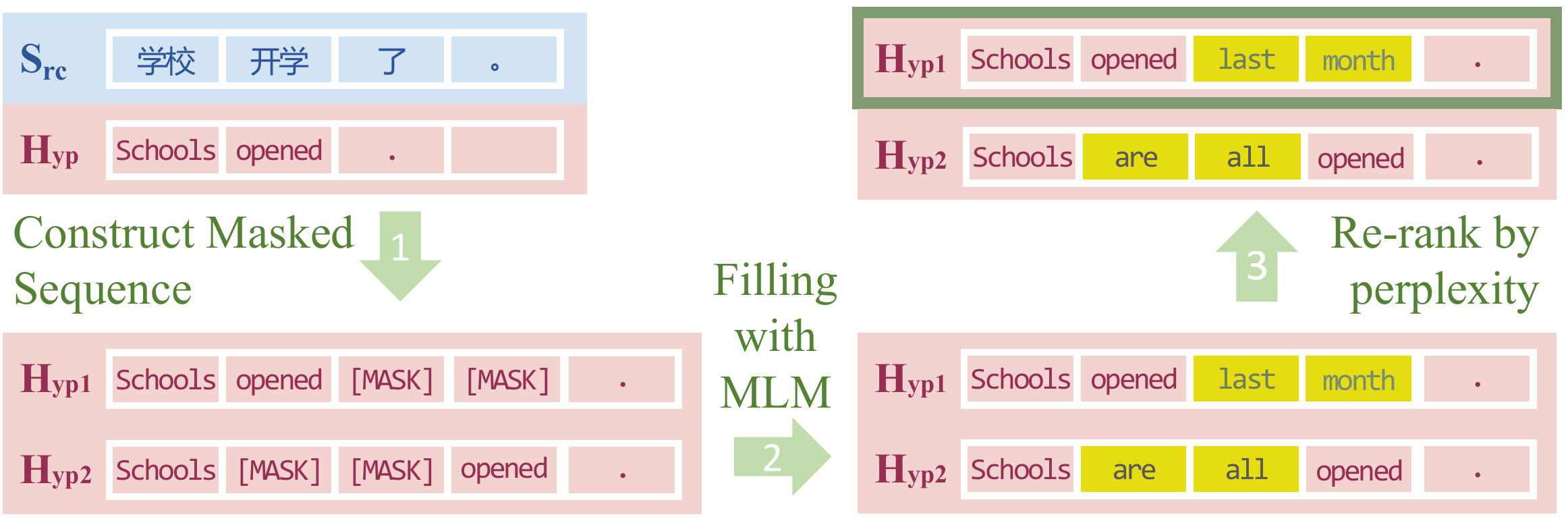}
    }
    \caption{Illustration of the pipeline of our synthetic data generated for Zh-En addition errors.}
    \label{figure.data}
\end{figure}

\begin{itemize}[noitemsep,nolistsep]
    \item Data filtering. To ensure that the constructed translation errors help model learning, the other words which are labeled as ``correctly translated'' should be highly aligned across languages. To achieve this, we use available sentence-level QE toolkits (\textit{e.g.},~\citealp[COMET-QE,][]{rei-etal-2020-comet};~\citealp[\textsc{UniTE},][]{wan-etal-2022-unite}) to help filter out low-quality examples.
    \item Mask-Filling. For each sentence pair, we first randomly choose the sentence to be processed (either $\mathbf{h}$ or $\mathbf{s}$).
    Then, we insert mask tokens into all available positions, and randomly determine the number of mask tokens to increase the diversity of synthetic samples.\footnote{\textcolor{Fix2}{Positions include the beginning and the end of sentence, as well as the spaces between any two adjacent words. The maximum number of consecutive mask tokens is set to 5 based on our empirical findings.}}
    After that, we fill the masked sequence with monolingual PLMs (\textit{e.g.}, \citealp[\textsc{BERT},][]{devlin2018bert}), apply recursive beam search~\cite{sellam-etal-2020-bleurt} to retain the fluency of generated samples, and construct the addition and omission errors in the \textsc{Hyp} and \textsc{Src} sentences, respectively.
    \item Reranking. After collecting filled candidate sequences, we use GPT-2~\cite{radford2019language} to collect their perplexity values for reranking.\footnote{English GPT-2: \href{https://huggingface.co/gpt2}{https://huggingface.co/gpt2}; Chinese GPT-2: \href{https://github.com/Morizeyao/GPT2-Chinese}{https://github.com/Morizeyao/GPT2-Chinese}; German GPT-2: \href{https://huggingface.co/dbmdz/german-gpt2}{https://huggingface.co/dbmdz/german-gpt2}.} Note that, different from direct selecting the top-1 candidate, we randomly choose one candidate among top-$k$ results to increase the diversity of translation errors.
\end{itemize}

Finally, we use WMT'14 En-De and WMT'17 Zh-En machine translation datasets -- two mainstream benchmarks to construct synthetic samples.
We collect 1.6M sentence pairs from these datasets and keep 0.3M high-quality pairs after filtering by QE toolkits.
\textcolor{Fix2}{To balance the diversity of synthetic dataset and the efficiency of data construction, in practice, we set the beam size to 8 during mask-filling, and select the top-8 samples when reranking.}
We totally collect 0.3M samples with well-labeled addition/omission errors
where each example contains \textsc{Hyp} $\mathbf{h}$, \textsc{Src} $\mathbf{s}$, and binary labels $\mathbf{y}$.

\paragraph{Dataset Relabeling}
\label{section.label}
Considering the available MQM dataset~\cite{freitag2021experts} is rather scarce for model training, we apply it as the test set to evaluate the performances of all models.
However, we find that the labels for some omission errors are on the target side (See Table~\ref{table_annotaion}), which brings disagreement for model evaluation~\cite{vamvas-sennrich-2022-little}.
To bridge such gap, we recruit two professional annotators to relabel the annotation of omission errors in the Zh-En direction.
We require them to tag those errors on the corresponding source side.\footnote{Details are discussed in Appendix \S\ref{appendix.human}.}
Totally, after combining the conventional and our relabeled annotations, we get 5,502 and 1,102 well-formatted examples for Zh-En and English-German (En-De), respectively.

\section{Experiments}
\label{section.exp}

\subsection{Experimental Settings}



\paragraph{Evaluation Setting}

Following~\newcite{vamvas-sennrich-2022-little}, we extract the examples containing \textit{Accuracy/Addition} and  \textit{Accuracy/Omission} errors in MQM datasets~\cite{freitag2021experts}.
As discussed in section \S\ref{section.label},  we replace the \textit{Accuracy/Omission} error in Chinese-English parts of the MQM'20 datasets with our relabeled ones.
During the evaluation, we re-weight each translation error word by the number of annotators tagging them, as the annotation is more confident if more annotators give the same annotation~\cite{monarch2021human}.
Notably, as the available PLMs would tokenize the sentence into subwords, we follow the related work \cite{ranasinghe2021}, treating the whole word as an error if any of its subwords is predicted as a translation error.

\paragraph{Baseline}
We include \textsc{Random}, \textsc{Contrastive Conditioning}~\cite{vamvas-sennrich-2022-little}, \textsc{Hallucination Detection}~\cite{zhou2021detecting}, \textsc{TransQuest}~\cite{ranasinghe2021}, \textsc{InfoXLM + DFL}~\cite{mahabadi2020end}, \textsc{InfoXLM + Fixed DFL}, \textsc{InfoXLM +GRL}~\cite{ganin2015unsupervised} as baselines.\footnote{For details about baselines and model settings, please refer to Appendix \S\ref{appendix.baseline} and \S\ref{appendix.model}.}.

\subsection{Main Results}
%
\begin{table*}[t]
    \centering
    \small
    \scalebox{0.75}{

        \begin{tabular}{llccccc}
            \toprule
            
            \multirow{2}{*}{\textbf{Line No.}} & \multirow{2}{*}{\textbf{Model}}  & \multicolumn{2}{c}{\textbf{Zh-En}} & \multicolumn{2}{c}{\textbf{En-De}} & \multirow{2}{*}{\textbf{Avg F1}}
            \\
            \cmidrule(l{3pt}r{3pt}){3-4}
            \cmidrule(l{3pt}r{3pt}){5-6}
             & & Addition & Omission & Addition & Omission &  \\
             \midrule
            \noalign{\vskip 0.0ex}
            \multicolumn{7}{c}{\textit{Baseline}} \\
            \cdashline{1-7}[1pt/2.5pt]\noalign{\vskip 0.5ex}
            (1) & \textsc{Random}                                                            & ~~1.1/55.0/~~2.2 & 14.6/50.4/22.6 & ~~4.9/61.0/~~9.1 & ~~0.6/55.0/~~1.2 & \colorbox{blue!10}{~~8.7} \\
            (2) & $^*$\textsc{Hallucination Detection}~\cite{zhou2021detecting}                     & ~~4.2/52.8/~~7.8 & -/-/- & -/-/-  & -/-/- & \colorbox{blue!10}{-} \\
            (3) & $^*$\textsc{TransQuest}~\cite{ranasinghe2021}                         & ~~1.5/25.7/~~2.8   & 14.6/92.2/25.2 & ~~1.7/42.2/~~3.3 & ~~0.1/100.0/~~1.2  & \colorbox{blue!10}{~~7.9} \\
            (4) & \textsc{Contrastive Conditioning}~\cite{vamvas-sennrich-2022-little}       \vspace{0.5ex}& 10.0/24.7/14.2   & 47.8/24.9/32.7 & 25.0/11.0/15.3   & 13.3/28.7/18.2   & \colorbox{blue!10}{20.1} \\

            \midrule
            \noalign{\vskip 0.0ex}
            \multicolumn{7}{c}{\textit{Trained by synthetic data}} \\
            \cdashline{1-7}[1pt/2.5pt]\noalign{\vskip 0.5ex}
            (5) & \textsc{mBART}                     & 17.7/~~4.6/~~7.3 & 39.8/28.8/33.4 & 17.8/28.4/21.9 & ~~6.7/13.4/~~9.0 & \colorbox{green!10}{17.9} \\
            (6) & \textsc{XLM-R}                & 17.5/~~2.9/~~5.0 & 41.6/30.9/35.5 & 40.5/29.9/34.4 & ~~5.6/28.3/~~9.3 & \colorbox{green!10}{21.0} \\
            (7) & \textsc{InfoXLM}             & 24.3/~~7.5/11.5 & 45.1/27.0/33.8 & 42.1/38.5/40.2 & ~~9.6/13.6/11.3 & \colorbox{green!10}{24.2} \\
            \midrule
            \noalign{\vskip 0.0ex}
            \multicolumn{7}{c}{\textit{Trained by synthetic data with shortcut learning reduction strategy}} \\
            \cdashline{1-7}[1pt/2.5pt]\noalign{\vskip 0.5ex}
            (8) & \textsc{XLM-R + GRL}~\cite{ganin2015unsupervised}  & 24.2/~~2.1/~~3.8 & 46.3/26.5/33.7 & 38.1/28.8/32.7 & ~~6.0/22.5/~~9.5 & \colorbox{red!10}{20.0} \\
            (9) & \textsc{InfoXLM + GRL}~\cite{ganin2015unsupervised}  & 24.5/~~3.3/~~5.8 & 50.1/25.2/33.6 & 42.4/40.0/41.2 & 10.9/18.6/13.8 & \colorbox{red!10}{23.6} \\
            (10) & \textsc{InfoXLM + DFL}~\cite{mahabadi2020end}          & ~~2.0/58.9/~~3.9   & 22.4/57.5/32.2 & 12.4/93.0/21.9   & ~~1.6/58.9/~~3.1 & \colorbox{red!10}{15.3} \\
            (11) & \textsc{InfoXLM + Fixed DFL}~\cite{mahabadi2020end}    & 38.8/~~3.7/15.5   & 56.2/23.5/33.1 & 46.4/19.2/27.2   & 20.9/15.5/17.8   & \colorbox{red!10}{23.4} \\

            (12) & \textsc{XLM-R + SLR (ours)}        & 33.5/~~6.1/10.3 & 44.1/34.4/38.7 & 38.3/39.6/38.9 & 13.2/58.1/\textbf{21.5} & \colorbox{red!10}{27.4} \\
            (13) & \textsc{InfoXLM + SLR (ours)}       & 27.5/13.5/\textbf{18.1} & 48.6/33.3/\textbf{39.5} & 37.5/47.6/\textbf{42.0} & 12.8/40.3/19.4 & \colorbox{red!10}{\textbf{29.8}} \\
            \bottomrule
             
        \end{tabular}
    }
    \caption{Precision/Recall/F1 scores of baselines and our methods. Best F1 scores are viewed in \textbf{bold}. Baselines marked with ``*'' indicate that the approaches are proposed for other tasks. Our model (\textsc{InfoXLM + SLR}) can achieve the best results than existing baselines and methods.}
    \vspace{-1.5em}
    \label{table_main}
\end{table*}

\subsubsection{FG-TED task}
Table \ref{table_main} shows the results for FG-TED task of all models. 
As seen, for all baselines, \textsc{Contrastive Conditioning}~\citep[Line 4,][]{vamvas-sennrich-2022-little} outperforms~\textsc{Hallucination Detection}~\citep[Line 2,][]{zhou2021detecting} and \textsc{TransQuest} methods~\citep[Line 3,][]{ranasinghe2021}.
Moreover, the average performance of \textsc{TransQuest} is even worse than \textsc{Random} (Line 1). 
This indicates that, those models which perform well on other related tasks may not be suitable for handling our proposed FG-TED task.

We first investigate the quality of our synthetic data.
As seen, applying different PLMs as backbones of our model can all surpass the performance of \textsc{TransQuest}.
Notably, the \textsc{XLM-R} approach (Line 6) and \textsc{TransQuest} baseline use the same model architecture.
This indicates the effectiveness of our synthetic data, which brings the improvement of 13.1 F1 scores on average.
Besides, when replacing the backbone with \textsc{InfoXLM} (Line 7), the performance further achieves 24.2, yielding an improvement of 3.2 F1 scores to the XLM-R approach.
The reason lies in the pre-training phase of \textsc{InfoXLM}: the PLM is enhanced with cross-lingual alignment semantics.
\textcolor{Fix2}{Besides, using \textsc{mBART}~\cite{liu2020multilingual} as the backbone of our model (Line 5) performs worst.
We think the reason is that, the encoder-decoder model architecture lacks the information interaction between two languages, failing to fully utilize the cross-lingual semantics for FG-TED~\cite{he2018layer}.}


\textcolor{Fix2}{Further, we compare our SRL with debiased focal loss~\citep[DFL,][]{mahabadi2020end} and gradient reversal layer~\citep[GRL,][]{ganin2015unsupervised}.}
As seen, our models can further improve their performances than those without SLR, revealing 6.4 (Line 12 vs. Line 6) and 5.6 (Line 13 vs. Line 7) averaged F1 scores using \textsc{XLM-R} and \textsc{InfoXLM} as backbones, respectively.
Especially for \textsc{InfoXLM + SLR} approach (Line 13), which delivers the highest averaged F1 scores at 29.8.
This reveals the effectiveness of our proposed SLR strategy, that the monolingual information in each sentence harms the FG-TED model learning.
By reducing the processing of monolingual features, the FG-TED models can deliver more accurate fine-grained error information.
Besides, compared to existing methods, introducing DFL or GRL to the model reveals a performance drop of averaged F1 at 8.9 and 0.6 (Line 11 vs. Line 7 and Line 9 vs. Line 7).
\textcolor{Fix2}{To investigate the reason why the performance of ``\textsc{InfoXLM} + DFL'' approach drops significantly, we apply a modified version ``\textsc{\textsc{InfoXLM} + Fixed DFL}''. See the analysis in Appendix \S\ref{appendix.dfl}.}

\subsubsection{Word-level QE \& CED task}
\textcolor{Fix2}{To illustrate the effectiveness of our methods on two sub-tasks of FG-TED, we conduct experiments on MLPE-QE dataset~\cite{specia-etal-2020-findings-wmt} for word-level QE task and ACES challenge set~\cite{amrhein2022aces} for CED task\footnote{\textcolor{Fix2}{To be consistent with their settings, for the former task, we conduct our experiments using the setting of the WMT21 word-level QE task~\cite{specia-etal-2020-findings-wmt} and ignore the ``[GAP]'' token; for the latter, we directly use the public repository for fine-tuning, and output a separate score.}}.}

\begin{table}[t]
    \centering
    \small
    \scalebox{0.8}{

        \begin{tabular}{lccccc}
            \toprule
            \multirow{2}{*}{\textbf{Model}} & \multicolumn{2}{c}{\textbf{En-De}} & \multicolumn{2}{c}{\textbf{En-Zh}} & \multirow{2}{*}{\textbf{Average}}\\  
           \cmidrule(l{3pt}r{3pt}){2-3}
           \cmidrule(l{3pt}r{3pt}){4-5}
            & \textbf{Source} & \textbf{Target} & \textbf{Source} &\textbf{Target} & \\
           \midrule
        \textsc{Baseline} & 32.3 & 37.0 & 24.1  & 24.7 & 29.5 \\
        \midrule
        \multicolumn{6}{c}{\textcolor{Fix2}{\textit{Backbone}: \textsc{XLM-R}}}  \\
        \cdashline{1-6}[1pt/2.5pt]\noalign{\vskip 0.5ex}         
        
        \textsc{+ FT} & 25.5 & 38.9  & 27.8 & 33.3 & 31.4 \\
        \textsc{+ FT*} & 25.4 & 38.2 & 28.2 & 34.5 & 31.6 \\
        \textsc{+ SYN + FT} & 26.7 & 41.3 & 29.5 & 36.5 & 33.5 \\
        \textsc{+ SYN + FT*} & \textbf{28.8} & \textbf{41.9} & 29.6 & \textbf{36.5} & \textbf{34.2} \\
        \textsc{+ SYN* + FT} & 25.6 & 41.8 & 30.0 & 35.6 & 33.3 \\
        \textsc{+ SYN* + FT*} & 28.5 & \textbf{41.9} & \textbf{30.1} & 36.1 & \textbf{34.2 }\\
        \midrule
        \multicolumn{6}{c}{\textcolor{Fix2}{\textit{Backbone}: \textsc{InfoXLM}}}  \\
        \cdashline{1-6}[1pt/2.5pt]\noalign{\vskip 0.5ex}         
        
        \textsc{+ FT} & 33.6 & 39.6 & 31.3 & 36.1 & 35.2 \\
        \textsc{+ FT*} & 33.4 & 40.1 & \textbf{31.4} & \textbf{36.4} & 35.3 \\
        \textsc{+ SYN + FT} & 33.9 & 39.9 & 30.3 & 35.1 & 34.8 \\
        \textsc{+ SYN + FT*} & \textbf{36.4} & 41.4 & 30.7 & 34.8 & 35.8\\
        \textsc{+ SYN* + FT} & 35.4 & \textbf{42.6} & 30.9 & 35.4 & \textbf{36.1} \\
        \textsc{+ SYN* + FT*} & 34.5 & 41.7 & 30.7 & 35.7 & 35.7\\
        \bottomrule
        \end{tabular}
    }
        \caption{Comparison on the Matthews correlation coefficient (MCC) on the word-level QE dataset for En-De and En-Zh. ``\textsc{Baseline}'' denotes the baseline results reported by WMT'21 organizers~\cite{specia-etal-2021-findings}. \textcolor{Fix2}{``+ FT'' and ``+ SYN'' denote the models trained on word-level QE training set and our synthetic data, respectively. ``*'' denotes we use the SLR strategy during the training phase. The best results for each backbone are in \textbf{bold}}.}
        \vspace{-1em}

    \label{table_word_level_qe}
\end{table}

Table~\ref{table_word_level_qe} exhibits the results of our methods on MLQE-PE dataset.\footnote{The baseline is trained on all En-XX directions while we only fine-tune our models on En-De and En-Zh directions, which leads to a performance gap in En-De source side.}
\textcolor{Fix2}{As seen:}
\textcolor{Fix2}{1) Directly applying the SLR strategy to the word-level-QE task always boost the performance (XLM-R + FT vs. XLM-R + FT*: 31.4 vs. 31.6 (+0.2); ~\textsc{InfoXLM} + FT vs~\textsc{InfoXLM} + FT*: 35.2 vs. 35.3 (+0.1)); 2) Simply using the synthetic data before fine-tuning get the improvement on all settings (+2.1 for XLM-R + FT, +2.6 for XLM-R + FT*, +0.5 for~\textsc{InfoXLM} + FT*) except slight drop at -0.4 for~\textsc{InfoXLM} + FT); 3) When applying the SLR strategy on the SYN phase,~\textsc{InfoXLM} + FT + SYN model performs better (34.8 vs. 36.1(+1.3)), and other scenarios demonstrate comparable results than baseline.
We conclude that, the combination of using synthetic data for continuous pre-training and applying the SLR strategy reveals a steady improvement in FG-TED model performance.}

\begin{table}[t]
    \centering
    \small
    \scalebox{0.8}{

        \begin{tabular}{lccc}
        \toprule
        
        \textbf{Model} & \textbf{Addition} & \textbf{Omission} & \textbf{Avg Score} \\ 
        \toprule
        \multicolumn{4}{c}{\textit{Baseline}} \\
        \cdashline{1-4}[1pt/2.5pt]\noalign{\vskip 0.5ex}         
        \textsc{WMT21-COMET-MQM} & -0.53 & 0.40 & -0.07 \\
        \textsc{WMT22-COMET-MQM} & 0.17 & 0.71 & 0.44 \\
        \textsc{UniTE-MQM} & -0.38 & 0.76 & 0.19 \\
        \midrule
        \multicolumn{4}{c}{\textcolor{Fix2}{\textit{Backbone}: \textsc{XLM-R}}}  \\
        \cdashline{1-4}[1pt/2.5pt]\noalign{\vskip 0.5ex}         
        \textsc{+ MQM} & -0.21 & 0.37 & 0.08 \\
        \textsc{+ SYN + MQM} & 0.28 & 0.75 & 0.52 \\
        \textsc{+ SYN* + MQM} & \textbf{0.30} & \textbf{0.76} & \textbf{0.53} \\
        \midrule
        \multicolumn{4}{c}{\textcolor{Fix2}{\textit{Backbone}: \textsc{InfoXLM}}}  \\
        \cdashline{1-4}[1pt/2.5pt]\noalign{\vskip 0.5ex} 
        \textsc{+ MQM} & -0.46 & 0.47 & 0.01 \\
        \textsc{+ SYN + MQM} & 0.18 & 0.83 & 0.51 \\
        \textsc{+ SYN* + MQM} & \textbf{0.33} & \textbf{0.86} & \textbf{0.60} \\
        
        \bottomrule
        \end{tabular}
    }
    
    \caption{\textcolor{Fix2}{Comparison on Kendall-tau correlation values on the ACES dataset. ``+ SYN'' and ``+ MQM'' means that the model is trained on the synthetic data and MQM datasets, respectively. ``*'' denotes we use the SLR strategy during the training phase.\protect\footnotemark~For baselines, we use official checkpoints for COMET~\cite{rei2022comet}, and we use the released checkpoint to fine-tune on MQM datasets for~\textsc{UniTE}~\cite{wan-etal-2022-unite}. The best results for each backbone are in \textbf{bold}. Note that, WMT22-COMET-MQM uses the multi-task strategy during the fine-tuning period, which is different from the others.}}
    \vspace{-1.5em}
   
    \label{table_aces}
\end{table}
\footnotetext{\textcolor{Fix2}{Our SLR serves the word-level classification tasks, which is inappropriate for fine-tuning on MQM, a benchmark that mainly serves for segment-level prediction.}}

Table \ref{table_aces} demonstrates the Kendall tau-like correlation results for addition and omission errors in the ACES datasets.
Our model outperforms the publicly released models (\citealp[COMET-21,][]{zerva-etal-2021-ist}; \citealp[WMT-COMET-22,][]{rei2022comet}; and \citealp[\textsc{UniTE},][]{wan-etal-2022-unite}).
\textcolor{Fix2}{Interestingly, training on synthetic data and applying the SLR strategy during training both boost the performances of our models on two backbones. This demonstrates that our SLR strategy and synthetic data are also useful to evaluate the translation quality of the hypothesis at the sentence level.}
\textcolor{Fix2}{In conclusion, our approach can not only handle FG-TED tasks, but is also suitable for its subtasks, \textit{i.e.}, word-level QE and CED.}

\section{Analysis}
\label{section.analysis}
We are interested to discover the characteristics of the FG-TED task and our method.
In this section, we direct our experiments with the following research questions:

\begin{itemize}[noitemsep,nolistsep]
    \item RQ1: How does \textcolor{Fix2}{the SLR} influence the utilization of monolingual features?
    \item RQ2: What is the performance of our method perform on the low-resource settings?
    \item RQ3: Whether our methods have the cross-lingual transferring ability or not?
    \item RQ4: What kind of pre-training method is the most effective for our FG-TED task?
\end{itemize}

\subsection{Influence of SLR (RQ1)}
\input{figure_logits}
We first explore the reason why \textcolor{Fix2}{SLR} helps our models on \textcolor{Fix2}{the} FG-TED task.
\textcolor{Fix2}{
Specially, if we modify the attention mask to remove the interactions between \textsc{Src} and \textsc{Hyp}, model predictions will be solely derived from the monolingual features.
All words in \textsc{Hyp} and \textsc{Src} should be predicted as addition and omission errors, respectively.}

Following this setting, we collect the results during inference to show the distribution of prediction probabilities, and compare the distributions of the predictions that are derived with \textsc{InfoXLM + SLR} and \textsc{InfoXLM} model (Exp 11 and 7 in Table~\ref{table_main}).
\textcolor{Fix2}{As Figure~\ref{fig.logits} illustrates, we can clearly observe that the model without SLR predicts almost all words as correctly translated ones.}
\textcolor{Fix2}{After introducing SLR into model training, our model can accurately identify most of the translation errors.}
\textcolor{Fix2}{These findings demonstrate that, based on the strong ability of PLM, the derived monolingual features suggest the FG-TED model mark all words as accurate ones, hardly utilizing cross-lingual information for translation error detection.}
\textcolor{Fix2}{After introducing SLR into model training, such influence can be moderated.}

\subsection{Generality on Low-Resource and Cross-Lingual Settings (RQ2, RQ3)}

Aside from table~\ref{table_main}, in this subsection, we further explore the generality of our model in low-resource and cross-lingual settings.


\paragraph{Low resource Setting} 
To further identify the generality of our model in the low-resource and cross-lingual settings, we first split the relabeled MQM dataset into the train, dev, and test sets with a ratio of 1:1:8, yielding 816, 888, and 6,885 examples for each subset.
To avoid the overlap of contextual information, we make sure that the examples assigned with the same \textsc{Src} sentence are involved in the same subset.
For the model fine-tuning, we randomly select 20, 40, 80, 400, and 816 examples to collect the model performances with different numbers of training examples.

\begin{figure}
    \centering
    \scalebox{1.0}{
        \begin{tikzpicture}
            \pgfplotsset{set layers}
            \pgfplotsset{every x tick label/.append style={font=\small}}
            \pgfplotsset{every y tick label/.append style={font=\small}}
            \begin{axis}[
                height=0.5 * \columnwidth,
                width=\columnwidth,
                xmin=0, xmax=825,
                ymin=5, ymax=40,
                xtick={40, 400, 816},
                xticklabels={$40$, $400$, $816$},
                ytick={5,25,40},
                yticklabels={$5$,$25$, $40$},
                ylabel={Avg F1},
                xlabel={Number of MQM Samples},
                ymajorgrids=true,
                grid style=dashed,
                legend cell align=left,
                legend style={
                    at={(0.5, 1.05)},
                    anchor=south,
                    font=\tiny,
            		legend columns=2},
            	every axis plot/.append style={thick},
                ]
            \addplot[
                color=blue,
                mark=o
                ]
                plot coordinates {
                    (20, 28.0)
                    (40, 32.2)
                    (80, 34.8)
                    (400, 38.3)
                    (817, 39.5)
                
            };
            \addlegendentry{\textsc{InfoXLM-SYN-MQM}}
            
            \addplot[
                color=green!80!black,
                mark=o
                ]
                plot coordinates {
                    (20, 17.4)
                    (40, 24.8)
                    (80, 32.1)
                    (400, 34.5)
                    (817, 36.8)
                };
            \addlegendentry{\textsc{XLM-R-SYN-MQM}}
            \addplot[
                color=black!80!white,
                mark=o
                ]
                plot coordinates {
                    (20, 8.8)
                    (40, 9.3)
                    (80, 10.5)
                    (400, 25.0)
                    (817, 28.7)
                
            };
            \addlegendentry{\textsc{InfoXLM-MQM}}
            \addplot[
                color=red!80!white,
                mark=*
                ]
                plot coordinates {
                    (0, 28.3)
                
            };
            \addlegendentry{\textsc{InfoXLM-SYN}}
            \end{axis}
        \end{tikzpicture}
        }
        \vspace{-2em}
        \caption{\textcolor{Fix2}{Comparison on low-resource setting. We use different scales of MQM dataset (x-axis) for training and collect the averaged model F1 scores (y-axis).} \textsc{InfoXLM} and \textsc{XLM-R}: the backbone we used. \textcolor{Fix2}{\textsc{syn}: model is first trained on synthetic examples.}}
        \label{figure.few_shot}
        \vspace{-2ex}

\end{figure}
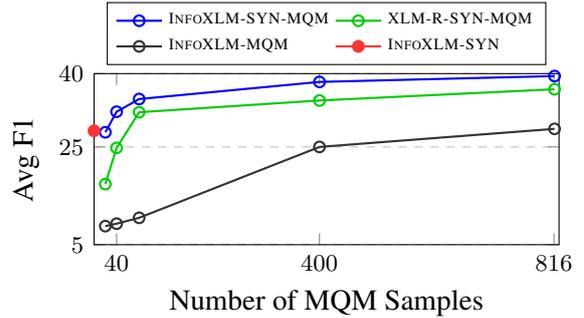

Figure~\ref{figure.few_shot} shows the result on the low-resource setting.
As seen, for the \textsc{InfoXLM-SYN-MQM} approach which is firstly trained with our synthetic data, fine-tuning on 20 MQM samples shows a comparable result with the model fine-tuned on the entire training set.
Besides, as the number of available MQM samples becomes larger for fine-tuning, the performance of the FG-TED model increases.
Meanwhile, using XLM-R as the backbone of our model (\textsc{XLM-R-SYN-MQM}) shows worse performance than \textsc{InfoXLM}, and the performance drop is consistent when using different numbers of training examples.

Building upon these findings, we claim that our synthetic data can significantly enhance the capability of the model.
In addition, the model with \textsc{InfoXLM} backbone outperforms that with \textsc{XLM-R}, especially when available fine-tuning data is extremely scarce (20 samples).
This may be related to the pre-training objective of those two PLMs.
Compared to XLM-R, the \textsc{InfoXLM} is enhanced with cross-lingual alignments information~\cite{chi-etal-2021-infoxlm}, which is important for our FG-TED task.

\paragraph{Cross-lingual Transferability} 
We also conduct experiments to explore the cross-lingual transferability of our model.
For this setting, the models are trained with the En-De annotated datasets and collect the predictions on Zh-En examples.

\begin{figure}[t]
    \centering
    \scalebox{1.0}{
        \begin{tikzpicture}
            \pgfplotsset{set layers}
            \pgfplotsset{every x tick label/.append style={font=\small}}
            \pgfplotsset{every y tick label/.append style={font=\small}}
            \begin{axis}[
                height=0.5 * \columnwidth,
                width=\columnwidth,
                xmin=0, xmax=120,
                ymin=5, ymax=50,
                xtick={3,  12,  24, 60, 120},
                xticklabels={$3$,  $12$, $24$, $60$, $120$},
                ytick={10,30,50},
                yticklabels={$10$,$30$, $50$},
                ylabel={Avg F1},
                xlabel={Number of En-De \textcolor{Fix2}{Synthetic} Samples (K)},
                ymajorgrids=true,
                grid style=dashed,
                legend cell align=left,
                legend style={
                    at={(0.5, 0.95)},
                    anchor=north,
                    font=\tiny,
            		legend columns=3},
            	every axis plot/.append style={thick},
                ]
            \addplot[
                color=blue,
                mark=o
                ]
                plot coordinates {
                    (0, 24.7)
                    (3, 28.3)
                    (12, 32.1)
                    (24, 35.7)
                    (60, 35.5)
                    (120, 35.5)
                
            };
            \addlegendentry{\textsc{InfoXLM}}
            
            \addplot[
                color=green!80!black,
                mark=o
                ]
                plot coordinates {
                    (0, 15.1)
                    (3, 16.1)
                    (12, 15.6)
                    (24, 16.2)
                    (60, 18.3)
                    (120, 19.1)
                };
            \addlegendentry{\textsc{XLM-R}}

            \addplot[
                color=black!80!white,
                mark=o
                ]
                plot coordinates {
                    (0, 12.7)
                    (120, 12.7)
                
            };
            \addlegendentry{\textsc{Random}}
                
            \end{axis}
        \end{tikzpicture}
        }
        \caption{\textcolor{Fix2}{Comparison on cross-lingual setting. We use different numbers of En-De synthetic samples (x-axis) for training, and collect the averaged performance on Zh-En test set (y-axis).}}
        \label{figure.zero_shot}

        \vspace{-3ex}

\end{figure}
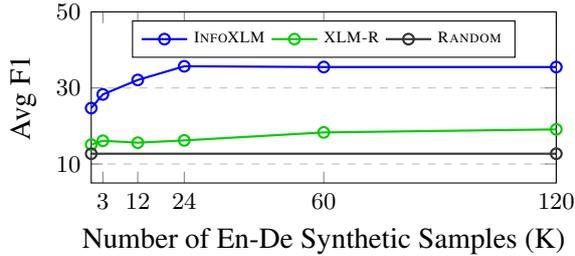

Figure~\ref{figure.zero_shot} shows the performance of models on the zero-shot setting.
As seen, when the number of synthetic data increases, the performances of two models with the backbones being \textsc{InfoXLM} and \textsc{XLM-R} increase, respectively.
This indicates that, our synthetic data can help models learn the core of the FG-TED task across languages.
Besides, the~\textsc{InfoXLM} approach shows consistently better than XLM-R, showing the importance of cross-lingual aligned semantics for our task.

In addition, for the cases where the number of synthetic examples is larger than 24k, both approaches show a limited improvement if more examples are used for training.

\subsection{Pre-training Objective (RQ4)} 

In this subsection, we further explore what training objective for PLMs helps our FG-TED task the most.
We collect the performances of PLMs trained with different training objectives on Zh-En subset and use 18.8M Zh-En parallel data (WMT'17 Zh-En Machine Translation benchmark) for pre-training using~\newcite{chi-etal-2021-infoxlm} repository.\footnote{https://github.com/microsoft/unilm/tree/master/infoxlm/src-infoxlm/infoxlm.}

As in Table~\ref{table.pre-train}, the PLM trained with MLM objective delivers the worst performance of FG-TED models.
Besides, the Translation Language Modeling~\citep[TLM,][]{conneau2019unsupervised} offers cross-lingual aligned semantics during training.
The corresponding backbone shows better results than that trained with MLM, indicating the importance of cross-lingual information on our task.

Besides, combining the MLM and TLM objectives shows a slight performance drop on average, and the implementation of \textsc{infoXLM},~\textit{i.e.}, combining MLM, TLM, and Cross-Lingual Contrast~\citep[XLCO,][]{chi-etal-2021-infoxlm}, further downgrades the performance of our FG-TED model.
Those results demonstrate that, the PLM which is pre-trained with cross-lingual aligned semantics can help the FG-TED model better identify addition and omission translation errors.
As~\newcite{chi-etal-2021-infoxlm} claims, TLM maximizes token-sequence mutual information, while XLCO maximizes the sentence-level mutual information between translation pairs.
Thus, TLM is more suitable for tasks that require fine-grained cross-lingual alignment information.

\begin{table}[t]
    \centering
    \scalebox{0.7}{
        \begin{tabular}{lccc}
           \toprule
           \multirow{2}{*}{\textbf{Model}}  & \multicolumn{2}{c}{\textbf{Zh-En}} &  \multirow{2}{*}{\textbf{Avg F1}}
            \\
            \cmidrule(l{3pt}r{3pt}){2-3}
             & Adddition & Omission  &  \\
             \midrule
            MLM              & ~~5.1/~~5.9/~~5.5     & 24.9/22.7/23.7 & 14.6 \\
            TLM              & 19.4/36.0/25.2     & 46.0/33.0/38.4 & 31.8 \\
            MLM + TLM        & 18.4/35.7/24.3     & 43.7/35.2/39.0 & 31.6 \\
            MLM + TLM + XLCO & 16.5/33.7/22.2    & 41.3/34.6/37.7 & 29.9 \\
            \bottomrule
        \end{tabular}
    }
    \caption{The performance of different pre-trianing objective on Zh-En part of MQM dataset.}
    \label{table.pre-train}
    \vspace{-1em}
\end{table}

\section{Conclusion and Future work}
\label{conclusion}

In this paper, we propose FG-TED task, which aims at delivering the type and position of translation errors.
We form our model following the word-level classification paradigm, and propose to reduce the shortcut learning of monolingual features raised by PLM.
Then, we utilize the mask-filling strategy with PLM to construct synthetic data for training, and relabel the misformatted examples of MQM datasets for reliable evaluation.
Experiments show that our methods can reduce the influence of shortcut learning, achieve promising performances on the relabeled dataset, and show high reliability on low-resource and cross-lingual settings.


Apart from addition and omission errors, mistranslation is also a type of critical error~\cite{freitag2021experts}. 
Specifically, it contains named entity errors, numerical errors, and terminology errors.
In our preliminary studies, we find that detecting the former two is quite easy for models trained on corresponding synthetic data.
While mistranslation errors are mainly raised in translating terminologies, especially for the cases where the terms are newly brought out in daily uses (e.g. the word "COVID" is rarely to be seen years early than 2020).
We believe that introducing external knowledge databases~\citep[\textit{e.g.},  multilingual knowledge graph,][]{yao2019kg,liu-EtAl:2022:WMT2} to our FG-TED models is expected to help identify the mistranslation errors~\cite{amrhein2022aces}, and we would like to leave this open problem to future work.

\bibliography{anthology}

\begin{thebibliography}{43}
\expandafter\ifx\csname natexlab\endcsname\relax\def\natexlab#1{#1}\fi

\bibitem[{Amrhein et~al.(2022)Amrhein, Moghe, and Guillou}]{amrhein2022aces}
Chantal Amrhein, Nikita Moghe, and Liane Guillou. 2022.
\newblock Aces: Translation accuracy challenge sets for evaluating machine
  translation metrics.
\newblock \emph{arXiv preprint arXiv:2210.15615}.

\bibitem[{Basu et~al.(2018)Basu, Pal, and Naskar}]{basu-etal-2018-keep}
Prasenjit Basu, Santanu Pal, and Sudip~Kumar Naskar. 2018.
\newblock \href {https://doi.org/10.18653/v1/W18-6457} {Keep it or not: Word
  level quality estimation for post-editing}.
\newblock In \emph{Proceedings of the Third Conference on Machine Translation:
  Shared Task Papers}, pages 759--764, Belgium, Brussels. Association for
  Computational Linguistics.

\bibitem[{Behnke et~al.(2022)Behnke, Fomicheva, and
  Specia}]{behnke-etal-2022-bias}
Hanna Behnke, Marina Fomicheva, and Lucia Specia. 2022.
\newblock \href {https://doi.org/10.18653/v1/2022.acl-long.104} {Bias
  mitigation in machine translation quality estimation}.
\newblock In \emph{Proceedings of the 60th Annual Meeting of the Association
  for Computational Linguistics (Volume 1: Long Papers)}, pages 1475--1487,
  Dublin, Ireland. Association for Computational Linguistics.

\bibitem[{Belinkov et~al.(2019)Belinkov, Poliak, Shieber, Van~Durme, and
  Rush}]{belinkov-etal-2019-adversarial}
Yonatan Belinkov, Adam Poliak, Stuart Shieber, Benjamin Van~Durme, and
  Alexander Rush. 2019.
\newblock \href {https://doi.org/10.18653/v1/S19-1028} {On adversarial removal
  of hypothesis-only bias in natural language inference}.
\newblock In \emph{Proceedings of the Eighth Joint Conference on Lexical and
  Computational Semantics (*{SEM} 2019)}, pages 256--262, Minneapolis,
  Minnesota. Association for Computational Linguistics.

\bibitem[{Chi et~al.(2021)Chi, Dong, Wei, Yang, Singhal, Wang, Song, Mao,
  Huang, and Zhou}]{chi-etal-2021-infoxlm}
Zewen Chi, Li~Dong, Furu Wei, Nan Yang, Saksham Singhal, Wenhui Wang, Xia Song,
  Xian-Ling Mao, Heyan Huang, and Ming Zhou. 2021.
\newblock \href {https://doi.org/10.18653/v1/2021.naacl-main.280} {{I}nfo{XLM}:
  An information-theoretic framework for cross-lingual language model
  pre-training}.
\newblock In \emph{Proceedings of the 2021 Conference of the North American
  Chapter of the Association for Computational Linguistics: Human Language
  Technologies}, pages 3576--3588, Online. Association for Computational
  Linguistics.

\bibitem[{Conneau et~al.(2019)Conneau, Khandelwal, Goyal, Chaudhary, Wenzek,
  Guzm{\'a}n, Grave, Ott, Zettlemoyer, and Stoyanov}]{conneau2019unsupervised}
Alexis Conneau, Kartikay Khandelwal, Naman Goyal, Vishrav Chaudhary, Guillaume
  Wenzek, Francisco Guzm{\'a}n, Edouard Grave, Myle Ott, Luke Zettlemoyer, and
  Veselin Stoyanov. 2019.
\newblock Unsupervised cross-lingual representation learning at scale.
\newblock \emph{arXiv preprint arXiv:1911.02116}.

\bibitem[{Conneau et~al.(2020)Conneau, Khandelwal, Goyal, Chaudhary, Wenzek,
  Guzm{\'{a}}n, Grave, Ott, Zettlemoyer, and
  Stoyanov}]{conneau2020unsupervised}
Alexis Conneau, Kartikay Khandelwal, Naman Goyal, Vishrav Chaudhary, Guillaume
  Wenzek, Francisco Guzm{\'{a}}n, Edouard Grave, Myle Ott, Luke Zettlemoyer,
  and Veselin Stoyanov. 2020.
\newblock {Unsupervised Cross-lingual Representation Learning at Scale}.
\newblock In \emph{Proceedings of the 58th Annual Meeting of the Association
  for Computational Linguistics (ACL)}.

\bibitem[{Devlin et~al.(2018)Devlin, Chang, Lee, and
  Toutanova}]{devlin2018bert}
Jacob Devlin, Ming-Wei Chang, Kenton Lee, and Kristina Toutanova. 2018.
\newblock Bert: Pre-training of deep bidirectional transformers for language
  understanding.
\newblock \emph{arXiv preprint arXiv:1810.04805}.

\bibitem[{Eyjolfsdottir et~al.(2017)Eyjolfsdottir, Branson, Yue, and
  Perona}]{eyjolfsdottir2017learning}
Eyrun Eyjolfsdottir, Kristin Branson, Yisong Yue, and Pietro Perona. 2017.
\newblock \href {https://openreview.net/forum?id=BkLhzHtlg} {Learning recurrent
  representations for hierarchical behavior modeling}.
\newblock In \emph{International Conference on Learning Representations}.

\bibitem[{Fomicheva et~al.(2022)Fomicheva, Sun, Fonseca, Zerva, Blain,
  Chaudhary, Guzm{\'a}n, Lopatina, Specia, and
  Martins}]{fomicheva-etal-2022-mlqe}
Marina Fomicheva, Shuo Sun, Erick Fonseca, Chrysoula Zerva, Fr{\'e}d{\'e}ric
  Blain, Vishrav Chaudhary, Francisco Guzm{\'a}n, Nina Lopatina, Lucia Specia,
  and Andr{\'e} F.~T. Martins. 2022.
\newblock \href {https://aclanthology.org/2022.lrec-1.530} {{MLQE}-{PE}: A
  multilingual quality estimation and post-editing dataset}.
\newblock In \emph{Proceedings of the Thirteenth Language Resources and
  Evaluation Conference}, pages 4963--4974, Marseille, France. European
  Language Resources Association.

\bibitem[{Fonseca et~al.(2019)Fonseca, Yankovskaya, Martins, Fishel, and
  Federmann}]{fonseca-etal-2019-findings}
Erick Fonseca, Lisa Yankovskaya, Andr{\'e} F.~T. Martins, Mark Fishel, and
  Christian Federmann. 2019.
\newblock \href {https://doi.org/10.18653/v1/W19-5401} {Findings of the {WMT}
  2019 shared tasks on quality estimation}.
\newblock In \emph{Proceedings of the Fourth Conference on Machine Translation
  (Volume 3: Shared Task Papers, Day 2)}, pages 1--10, Florence, Italy.
  Association for Computational Linguistics.

\bibitem[{Freitag et~al.(2021{\natexlab{a}})Freitag, Foster, Grangier,
  Ratnakar, Tan, and Macherey}]{freitag2021experts}
Markus Freitag, George Foster, David Grangier, Viresh Ratnakar, Qijun Tan, and
  Wolfgang Macherey. 2021{\natexlab{a}}.
\newblock Experts, errors, and context: A large-scale study of human evaluation
  for machine translation.
\newblock \emph{Transactions of the Association for Computational Linguistics},
  9:1460--1474.

\bibitem[{Freitag et~al.(2021{\natexlab{b}})Freitag, Rei, Mathur, Lo, Stewart,
  Foster, Lavie, and Bojar}]{freitag-etal-2021-results}
Markus Freitag, Ricardo Rei, Nitika Mathur, Chi-kiu Lo, Craig Stewart, George
  Foster, Alon Lavie, and Ond{\v{r}}ej Bojar. 2021{\natexlab{b}}.
\newblock \href {https://aclanthology.org/2021.wmt-1.73} {Results of the
  {WMT}21 metrics shared task: Evaluating metrics with expert-based human
  evaluations on {TED} and news domain}.
\newblock In \emph{Proceedings of the Sixth Conference on Machine Translation},
  pages 733--774, Online. Association for Computational Linguistics.

\bibitem[{Ganin and Lempitsky(2015)}]{ganin2015unsupervised}
Yaroslav Ganin and Victor Lempitsky. 2015.
\newblock Unsupervised domain adaptation by backpropagation.
\newblock In \emph{International conference on machine learning}, pages
  1180--1189. PMLR.

\bibitem[{Ganin et~al.(2016)Ganin, Ustinova, Ajakan, Germain, Larochelle,
  Laviolette, Marchand, and Lempitsky}]{ganin2016domain}
Yaroslav Ganin, Evgeniya Ustinova, Hana Ajakan, Pascal Germain, Hugo
  Larochelle, Fran{\c{c}}ois Laviolette, Mario Marchand, and Victor Lempitsky.
  2016.
\newblock Domain-adversarial training of neural networks.
\newblock \emph{The journal of machine learning research}, 17(1):2096--2030.

\bibitem[{Gao et~al.(2020)Gao, Wang, Herold, Yang, and
  Ney}]{gao-etal-2020-towards}
Yingbo Gao, Weiyue Wang, Christian Herold, Zijian Yang, and Hermann Ney. 2020.
\newblock \href {https://aclanthology.org/2020.aacl-main.25} {Towards a better
  understanding of label smoothing in neural machine translation}.
\newblock In \emph{Proceedings of the 1st Conference of the Asia-Pacific
  Chapter of the Association for Computational Linguistics and the 10th
  International Joint Conference on Natural Language Processing}, pages
  212--223, Suzhou, China. Association for Computational Linguistics.

\bibitem[{He et~al.(2018)He, Tan, Xia, He, Qin, Chen, and Liu}]{he2018layer}
Tianyu He, Xu~Tan, Yingce Xia, Di~He, Tao Qin, Zhibo Chen, and Tie-Yan Liu.
  2018.
\newblock Layer-wise coordination between encoder and decoder for neural
  machine translation.
\newblock \emph{Advances in Neural Information Processing Systems}, 31.

\bibitem[{Kepler et~al.(2019)Kepler, Tr{\'e}nous, Treviso, Vera, and
  Martins}]{kepler-etal-2019-openkiwi}
Fabio Kepler, Jonay Tr{\'e}nous, Marcos Treviso, Miguel Vera, and Andr{\'e}
  F.~T. Martins. 2019.
\newblock \href {https://doi.org/10.18653/v1/P19-3020} {{O}pen{K}iwi: An open
  source framework for quality estimation}.
\newblock In \emph{Proceedings of the 57th Annual Meeting of the Association
  for Computational Linguistics: System Demonstrations}, pages 117--122,
  Florence, Italy. Association for Computational Linguistics.

\bibitem[{Kim et~al.(2017)Kim, Lee, and Na}]{kim-etal-2017-predictor}
Hyun Kim, Jong-Hyeok Lee, and Seung-Hoon Na. 2017.
\newblock \href {https://doi.org/10.18653/v1/W17-4763} {Predictor-estimator
  using multilevel task learning with stack propagation for neural quality
  estimation}.
\newblock In \emph{Proceedings of the Second Conference on Machine
  Translation}, pages 562--568, Copenhagen, Denmark. Association for
  Computational Linguistics.

\bibitem[{Lewis et~al.(2019)Lewis, Liu, Goyal, Ghazvininejad, Mohamed, Levy,
  Stoyanov, and Zettlemoyer}]{lewis2019bart}
Mike Lewis, Yinhan Liu, Naman Goyal, Marjan Ghazvininejad, Abdelrahman Mohamed,
  Omer Levy, Ves Stoyanov, and Luke Zettlemoyer. 2019.
\newblock Bart: Denoising sequence-to-sequence pre-training for natural
  language generation, translation, and comprehension.
\newblock \emph{arXiv preprint arXiv:1910.13461}.

\bibitem[{Lin et~al.(2017)Lin, Goyal, Girshick, He, and
  Doll{\'a}r}]{lin2017focal}
Tsung-Yi Lin, Priya Goyal, Ross Girshick, Kaiming He, and Piotr Doll{\'a}r.
  2017.
\newblock Focal loss for dense object detection.
\newblock In \emph{Proceedings of the IEEE international conference on computer
  vision}, pages 2980--2988.

\bibitem[{Liu et~al.(2022)Liu, Qiao, Wu, Chang, Zhang, Zhao, Peng, tao, Yang,
  Qin, Guo, Wang, Li, Li, and Zhao}]{liu-EtAl:2022:WMT2}
Yilun Liu, Xiaosong Qiao, Zhanglin Wu, Su~Chang, Min Zhang, Yanqing Zhao, Song
  Peng, shimin tao, Hao Yang, Ying Qin, Jiaxin Guo, Minghan Wang, Yinglu Li,
  Peng Li, and Xiaofeng Zhao. 2022.
\newblock \href {https://aclanthology.org/2022.wmt-1.48} {Partial could be
  better than whole. hw-tsc 2022 submission for the metrics shared task}.
\newblock In \emph{Proceedings of the Seventh Conference on Machine
  Translation}, pages 549--557, Abu Dhabi. Association for Computational
  Linguistics.

\bibitem[{Liu et~al.(2020)Liu, Gu, Goyal, Li, Edunov, Ghazvininejad, Lewis, and
  Zettlemoyer}]{liu2020multilingual}
Yinhan Liu, Jiatao Gu, Naman Goyal, Xian Li, Sergey Edunov, Marjan
  Ghazvininejad, Mike Lewis, and Luke Zettlemoyer. 2020.
\newblock Multilingual denoising pre-training for neural machine translation.
\newblock \emph{Transactions of the Association for Computational Linguistics},
  8:726--742.

\bibitem[{Mahabadi et~al.(2020)Mahabadi, Belinkov, and
  Henderson}]{mahabadi2020end}
Rabeeh~Karimi Mahabadi, Yonatan Belinkov, and James Henderson. 2020.
\newblock End-to-end bias mitigation by modelling biases in corpora.
\newblock In \emph{Proceedings of the 58th Annual Meeting of the Association
  for Computational Linguistics}, pages 8706--8716.

\bibitem[{Monarch(2021)}]{monarch2021human}
Robert~Munro Monarch. 2021.
\newblock \emph{Human-in-the-Loop Machine Learning: Active learning and
  annotation for human-centered AI}.
\newblock Simon and Schuster.

\bibitem[{Pereyra et~al.(2017)Pereyra, Tucker, Chorowski, Kaiser, and
  Hinton}]{pereyra-etal-2017-regularizing}
Gabriel Pereyra, George Tucker, Jan Chorowski, Lukasz Kaiser, and Geoffrey~E.
  Hinton. 2017.
\newblock \href {https://openreview.net/forum?id=HyhbYrGYe} {Regularizing
  neural networks by penalizing confident output distributions}.
\newblock In \emph{5th International Conference on Learning Representations,
  {ICLR} 2017, Toulon, France, April 24-26, 2017, Workshop Track Proceedings}.
  OpenReview.net.

\bibitem[{Radford et~al.(2019)Radford, Wu, Child, Luan, Amodei, Sutskever
  et~al.}]{radford2019language}
Alec Radford, Jeffrey Wu, Rewon Child, David Luan, Dario Amodei, Ilya
  Sutskever, et~al. 2019.
\newblock Language models are unsupervised multitask learners.
\newblock \emph{OpenAI blog}, 1(8):9.

\bibitem[{Ranasinghe et~al.(2021)Ranasinghe, Orasan, and
  Mitkov}]{ranasinghe2021}
Tharindu Ranasinghe, Constantin Orasan, and Ruslan Mitkov. 2021.
\newblock An exploratory analysis of multilingual word level quality estimation
  with cross-lingual transformers.
\newblock In \emph{Proceedings of the 59th Annual Meeting of the Association
  for Computational Linguistics}.

\bibitem[{Rei et~al.(2022)Rei, de~Souza, Alves, Zerva, Farinha, Glushkova,
  Lavie, Coheur, and Martins}]{rei2022comet}
Ricardo Rei, Jos{\'e}~GC de~Souza, Duarte Alves, Chrysoula Zerva, Ana~C
  Farinha, Taisiya Glushkova, Alon Lavie, Luisa Coheur, and Andr{\'e}~FT
  Martins. 2022.
\newblock Comet-22: Unbabel-ist 2022 submission for the metrics shared task.
\newblock In \emph{Proceedings of the Seventh Conference on Machine
  Translation, Abu Dhabi. Association for Computational Linguistics}.

\bibitem[{Rei et~al.(2020)Rei, Stewart, Farinha, and
  Lavie}]{rei-etal-2020-comet}
Ricardo Rei, Craig Stewart, Ana~C Farinha, and Alon Lavie. 2020.
\newblock \href {https://doi.org/10.18653/v1/2020.emnlp-main.213} {{COMET}: A
  neural framework for {MT} evaluation}.
\newblock In \emph{Proceedings of the 2020 Conference on Empirical Methods in
  Natural Language Processing (EMNLP)}, pages 2685--2702, Online. Association
  for Computational Linguistics.

\bibitem[{Sellam et~al.(2020)Sellam, Das, and Parikh}]{sellam-etal-2020-bleurt}
Thibault Sellam, Dipanjan Das, and Ankur Parikh. 2020.
\newblock \href {https://doi.org/10.18653/v1/2020.acl-main.704} {{BLEURT}:
  Learning robust metrics for text generation}.
\newblock In \emph{Proceedings of the 58th Annual Meeting of the Association
  for Computational Linguistics}, pages 7881--7892, Online. Association for
  Computational Linguistics.

\bibitem[{Specia et~al.(2020)Specia, Blain, Fomicheva, Fonseca, Chaudhary,
  Guzm{\'a}n, and Martins}]{specia-etal-2020-findings-wmt}
Lucia Specia, Fr{\'e}d{\'e}ric Blain, Marina Fomicheva, Erick Fonseca, Vishrav
  Chaudhary, Francisco Guzm{\'a}n, and Andr{\'e} F.~T. Martins. 2020.
\newblock \href {https://aclanthology.org/2020.wmt-1.79} {Findings of the {WMT}
  2020 shared task on quality estimation}.
\newblock In \emph{Proceedings of the Fifth Conference on Machine Translation},
  pages 743--764, Online. Association for Computational Linguistics.

\bibitem[{Specia et~al.(2021)Specia, Blain, Fomicheva, Zerva, Li, Chaudhary,
  and Martins}]{specia-etal-2021-findings}
Lucia Specia, Fr{\'e}d{\'e}ric Blain, Marina Fomicheva, Chrysoula Zerva,
  Zhenhao Li, Vishrav Chaudhary, and Andr{\'e} F.~T. Martins. 2021.
\newblock \href {https://aclanthology.org/2021.wmt-1.71} {Findings of the {WMT}
  2021 shared task on quality estimation}.
\newblock In \emph{Proceedings of the Sixth Conference on Machine Translation},
  pages 684--725, Online. Association for Computational Linguistics.

\bibitem[{Specia et~al.(2018)Specia, Blain, Logacheva, F.~Astudillo, and
  Martins}]{specia-etal-2018-findings}
Lucia Specia, Fr{\'e}d{\'e}ric Blain, Varvara Logacheva, Ram{\'o}n
  F.~Astudillo, and Andr{\'e} F.~T. Martins. 2018.
\newblock \href {https://doi.org/10.18653/v1/W18-6451} {Findings of the {WMT}
  2018 shared task on quality estimation}.
\newblock In \emph{Proceedings of the Third Conference on Machine Translation:
  Shared Task Papers}, pages 689--709, Belgium, Brussels. Association for
  Computational Linguistics.

\bibitem[{Vamvas and Sennrich(2021)}]{vamvas-sennrich-2021-contrastive}
Jannis Vamvas and Rico Sennrich. 2021.
\newblock \href {https://doi.org/10.18653/v1/2021.emnlp-main.803} {Contrastive
  conditioning for assessing disambiguation in {MT}: {A} case study of
  distilled bias}.
\newblock In \emph{Proceedings of the 2021 Conference on Empirical Methods in
  Natural Language Processing}, pages 10246--10265, Online and Punta Cana,
  Dominican Republic. Association for Computational Linguistics.

\bibitem[{Vamvas and Sennrich(2022)}]{vamvas-sennrich-2022-little}
Jannis Vamvas and Rico Sennrich. 2022.
\newblock \href {https://doi.org/10.18653/v1/2022.acl-short.53} {As little as
  possible, as much as necessary: Detecting over- and undertranslations with
  contrastive conditioning}.
\newblock In \emph{Proceedings of the 60th Annual Meeting of the Association
  for Computational Linguistics (Volume 2: Short Papers)}, pages 490--500,
  Dublin, Ireland. Association for Computational Linguistics.

\bibitem[{Wan et~al.(2022)Wan, Liu, Yang, Zhang, Chen, Wong, and
  Chao}]{wan-etal-2022-unite}
Yu~Wan, Dayiheng Liu, Baosong Yang, Haibo Zhang, Boxing Chen, Derek Wong, and
  Lidia Chao. 2022.
\newblock \href {https://doi.org/10.18653/v1/2022.acl-long.558} {{U}ni{TE}:
  Unified translation evaluation}.
\newblock In \emph{Proceedings of the 60th Annual Meeting of the Association
  for Computational Linguistics (Volume 1: Long Papers)}, pages 8117--8127,
  Dublin, Ireland. Association for Computational Linguistics.

\bibitem[{Xu et~al.(2022)Xu, Tuan, Lu, Saxon, Li, and Wang}]{xu2022not}
Wenda Xu, Yilin Tuan, Yujie Lu, Michael Saxon, Lei Li, and William~Yang Wang.
  2022.
\newblock Not all errors are equal: Learning text generation metrics using
  stratified error synthesis.
\newblock \emph{arXiv preprint arXiv:2210.05035}.

\bibitem[{Yao et~al.(2019)Yao, Mao, and Luo}]{yao2019kg}
Liang Yao, Chengsheng Mao, and Yuan Luo. 2019.
\newblock Kg-bert: Bert for knowledge graph completion.
\newblock \emph{arXiv preprint arXiv:1909.03193}.

\bibitem[{Zerva et~al.(2021)Zerva, van Stigt, Rei, Farinha, Ramos, C.~de Souza,
  Glushkova, Vera, Kepler, and Martins}]{zerva-etal-2021-ist}
Chrysoula Zerva, Daan van Stigt, Ricardo Rei, Ana~C Farinha, Pedro Ramos,
  Jos{\'e}~G. C.~de Souza, Taisiya Glushkova, Miguel Vera, Fabio Kepler, and
  Andr{\'e} F.~T. Martins. 2021.
\newblock \href {https://aclanthology.org/2021.wmt-1.102} {{IST}-unbabel 2021
  submission for the quality estimation shared task}.
\newblock In \emph{Proceedings of the Sixth Conference on Machine Translation},
  pages 961--972, Online. Association for Computational Linguistics.

\bibitem[{Zhang et~al.(2021)Zhang, Yang, and Zhao}]{zhang2021retrospective}
Zhuosheng Zhang, Junjie Yang, and Hai Zhao. 2021.
\newblock Retrospective reader for machine reading comprehension.
\newblock In \emph{Thirty-Fifth {AAAI} Conference on Artificial Intelligence,
  {AAAI} 2021, Thirty-Third Conference on Innovative Applications of Artificial
  Intelligence, {IAAI} 2021, The Eleventh Symposium on Educational Advances in
  Artificial Intelligence, {EAAI} 2021, Virtual Event, February 2-9, 2021},
  pages 14506--14514.

\bibitem[{Zhong et~al.(2022)Zhong, Liu, Yin, Mao, Jiao, Liu, Zhu, Ji, and
  Han}]{zhong2022towards}
Ming Zhong, Yang Liu, Da~Yin, Yuning Mao, Yizhu Jiao, Pengfei Liu, Chenguang
  Zhu, Heng Ji, and Jiawei Han. 2022.
\newblock Towards a unified multi-dimensional evaluator for text generation.
\newblock \emph{arXiv preprint arXiv:2210.07197}.

\bibitem[{Zhou et~al.(2021)Zhou, Neubig, Gu, Diab, Guzm{\'a}n, Zettlemoyer, and
  Ghazvininejad}]{zhou2021detecting}
Chunting Zhou, Graham Neubig, Jiatao Gu, Mona Diab, Francisco Guzm{\'a}n, Luke
  Zettlemoyer, and Marjan Ghazvininejad. 2021.
\newblock Detecting hallucinated content in conditional neural sequence
  generation.
\newblock In \emph{Findings of the Association for Computational Linguistics:
  ACL-IJCNLP 2021}, pages 1393--1404.

\end{thebibliography}

\appendix
\section{Ethics Statement}
In this paper, we propose FG-TED which aims at detecting both error types and error positions in translation hypothesis.
Our method does not involve any ethical issues, and our annotation process respects human rights and provides adequate wages in line with local wage levels.
However, the datasets as well as the PLMs used in this research may have some negative impacts, such as gender and social bias.
Sufferring from these issues is inevitable for us.
We suggest that users should be aware of potential risks for their own purposes.

\section{Limitations}
\label{appendix.limitations}
We list the two main limitations of this work as follows:

\begin{itemize}
    \item  Low resource language pairs. Although our experiments have shown great results in Zh-En and En-De directions, our methods rely on the PLM's strong representation ability in high-resource languages. We have not validated our approach to low-resource languages.
    \item  Sufficient amount of data. In our research, all our experiments are conducted on MQM dataset which does not have enough data to support a whole supervised training period. We did not discuss the performance of our methods when there have a sufficient amount of labeled data.
\end{itemize}

\section{Human Evaluation}
\label{appendix.human}
\begin{table*}[t]
    \centering
    \scalebox{0.9}{
        \begin{tabular}{lcccccc}
        \toprule
        & \multicolumn{3}{c}{segment-level} &  \multicolumn{3}{c}{word-level} \\
        \cmidrule(l{2pt}r{2pt}){2-4}\cmidrule(l{2pt}r{2pt}){5-7}
         & \textbf{Zh-En} & \textbf{En-De} & \textbf{Total}    & \textbf{Zh-En} & \textbf{En-De} & \textbf{Total}  \\
         \midrule
         \textbf{Addition} & 1166 & 1055 & 2201 & 3071 & 2088 & 5199 \\
         \textbf{Omission} & 4622 & 65 & 4687 & 30516 & 258 & 30414 \\
         \textbf{Total} & 5502 & 1120 & 6622 & 33227 & 2344 & 35571 \\
         
         \bottomrule
        \end{tabular}
    }
    \caption{Statistics of our datasets.}
    \label{table_statistic}
\end{table*}

We initially established a set of criteria for evaluating, which includes the context of the task, a thorough description, and illustrations of annotation. 
In detail, annotators are asked to perform the following three steps in sequence.
\begin{itemize}
    \item Comparing the difference between existing MQM annotations and the original translation hypothesis.
    \item Determine whether the current MQM supplementary results for omission errors have introduced other errors, such as typo errors and repeat inputs.
    \item Labeling the corresponding omission errors on the source side while ignoring the bias introduced by MQM annotations.
\end{itemize}
Then we set an entry barrier for annotators.
Specifically, we set up a training program and a 100-example preliminary annotating exam for each annotator.
We do not recruit annotators whose annotation accuracy rate is less than 95\% in ths exam.
Totally, we pay 0.8\$ for each annotator per segement in average.
Exampls are shown in the second example in Table. \ref{table_mqm}.

\paragraph{Inter-annotator agreement} To ensure the quality of our annotations, we randomly sample 10\% of the data and recruit another professional annotator to label them again without seeing the existing annotations. 
We use Extract-Match (EM) to measure our annotator's reliability.\footnote{EM is commonly used in span prediction tasks \cite{zhang2021retrospective, eyjolfsdottir2017learning}.}
The result is 0.97 among our annotators, which indicates that our relabeled dataset is labeled accurately.

\section{Fixed Debiased Focal Loss (\textsc{Fixed DFL})}
\label{appendix.dfl}
We think the reason why DFL performs worse lies in the debiasing strategy, where DFL is designed to mainly address data-level biases.
For Nature Language Inference (NLI) tasks, the negation words like ``not'' are strongly linked with the contradiction label~\cite{mahabadi2020end}.
To solve this problem, the DFL strategy utilizes a biased model to help reweight those examples containing negation words, and avoid the main model from overfitting them.
However, we find that \textsc{InfoXLM + DFL} method (Line 8 in Table \ref{table_main}) shows high recall and low precision in all directions.
We think the reason lies in that, the monolingual features are often correlated with the ``good'' label because the words labeled with ``good'' are generally fluent and grammatically correct.
Thus, the biased model learns to hinder the base model from learning on ``good'' labels and lead to learning collapse.
To overcome this phenomenon, we fix it by removing the loss reweighting mechanism for correctly translated words.
As seen, the Fixed DFL (Line 9 in Table \ref{table_main}) achieves 23.4 F1 scores on average, 8.1 scores higher than the conventional DFL strategy (Line 8 in Table \ref{table_main}).
This verifies our thought, that the conventional DFL approach harms the shortcut learning reduction of FG-TED model.

\section{Baseline}
\label{appendix.baseline}
For comparison, we involve the following baselines for comparison:
\begin{itemize}[noitemsep,nolistsep]
    \item \textsc{Random}: \textcolor{Fix2}{To make the comparison more feasible, we expect to involve a lower bound of FG-TED task on MQM benchmark. In detail, we randomly tag the prediction of each token as a ``correct'' one with a chance of 50\%.}
    \item \textsc{Contrastive-Conditioning}~\cite{vamvas-sennrich-2022-little}\footnote{https://github.com/ZurichNLP/coverage-contrastive-conditioning}: It applies contrastive conditioning with \textsc{mBART} to detect addition and omission errors.
    \item \textsc{Hallucination Detection~\cite{zhou2021detecting}\footnote{https://github.com/violet-zct/fairseq-detect-hallucination}}: It uses \textsc{BART}~\cite{lewis2019bart} to generate hallucinated translations, and detect word-level hallucination errors. In our experiment, we treat the hallucination errors as addition errors, and directly use the officially released checkpoint for comparison.
    \item \textsc{TransQuest}~\cite{ranasinghe2021}\footnote{https://github.com/TharinduDR/TransQuest}: It is a strong baseline in word-level QE task, which uses \textsc{XLM-R}~\cite{conneau2019unsupervised} and train on the supervised dataset. We use the released model for performance comparison.
    \item \textsc{InfoXLM + DFL}~\cite{mahabadi2020end}: After defining the debiased focal loss (DFL), it trains an additional debiasing model to help the training phase of a classification model. In this work, we use the DFL to see whether it can reduce the influence of shortcut learning.
    \item \textsc{InfoXLM + Fixed DFL}: We remove the reweighting mechanism of the debiased model in \textsc{InfoXLM + DFL} approach for stable training.
    \item \textsc{InfoXLM + GRL}~\cite{ganin2015unsupervised}: we adapt the gradient reversal approach into our models which use an adversarial loss to avoid shortcut learning.
\end{itemize}

\section{Experiments setting}
\label{appendix.model}
\paragraph{Recursive beam search}  We perform multiple iterations (depends on the number of ``[mask]'' tokens) to complete thas mask-filling process. In each iteration, we consider all feasible positions and only fill a ``[mask]''. To acheive this, we select top-8 (beam size) tokens based on the probabilities given by LMs and substitute the corresponding ``[mask]'' with the selected token. Subsequently, the updated text is passed on the next iteration. It is important to note that the masks are not filled in a sequential manner, such as from left to right or right to left.
\paragraph{Model setting} We develop our method based on the COMET repository~\cite{rei-etal-2020-comet},\footnote{https://github.com/Unbabel/COMET} and use \textsc{InfoXLM}~\cite{chi-etal-2021-infoxlm}\footnote{https://huggingface.co/microsoft/infoxlm-large} as the backbone of our model. 
The classifier is a three-layer feedforward network, whose output dimensionalities of all layers are 3,072, 1,024, and 2, respectively.
Besides, the activation between any two adjacent linear layers is hyperbolic tangent.
During training, we randomly select 2,000 synthetic samples as the development set.
The learning rates for the PLM and the classifier are $1.0 \times 10^{-5}$ and $1.0 \times 10^{-4}$, respectively.
Considering the error words in datasets are more scarce than correctly translated ones, during each learning step, we randomly detached the 90\% correctly translated words from gradient back-propagation to balance the model training.
For all backbones, we use the large setting to make them comparable.
All experiments are conducted on one single Nvidia V100 (32GB) GPU device.

\section{Hyper-parameter $\alpha$}
\label{appendix.parameter}
\subsection{Language-Pair Agnostic}
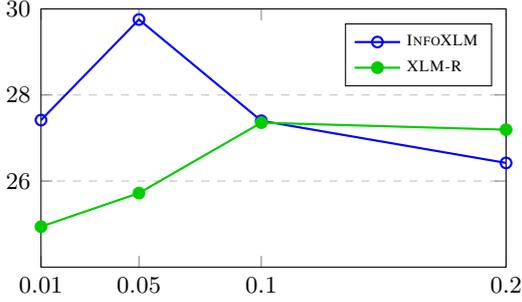
\begin{figure}
    \centering
        \begin{tikzpicture}
            \pgfplotsset{set layers}
            \pgfplotsset{every x tick label/.append style={font=\small}}
            \pgfplotsset{every y tick label/.append style={font=\small}}
            \begin{axis}[
                height=0.65 * \columnwidth,
                width=\columnwidth,
                xmin=0.01, xmax=0.2,
                ymin=24, ymax=30,
                xtick={0.01, 0.05, 0.1, 0.2},
                xticklabels={$0.01$, $0.05$, $0.1$, $0.2$},
                ytick={26,28,30},
                yticklabels={$26$, $28$, $30$},
                ymajorgrids=true,
                grid style=dashed,
                legend cell align=left,
                legend style={
                    at={(0.97, 0.70)},
                    anchor=south east,
                    font=\tiny,
            		legend columns=1},
            	every axis plot/.append style={thick},
                ]
            \addplot[
                color=blue,
                mark=o
                ]
                plot coordinates {
                    (0.01, 27.4164)
                    (0.05, 29.7526)
                    (0.1, 27.4000)
                    (0.2, 26.4208)
            };
            \addlegendentry{\textsc{InfoXLM}}
            
            \addplot[
                color=green!80!black,
                mark=*
                ]
                plot coordinates {
                    (0.01, 24.9402)
                    (0.05, 25.7199)
                    (0.1, 27.3558)
                    (0.2, 27.1934)
                };
            \addlegendentry{\textsc{XLM-R}}
            \end{axis}
        \end{tikzpicture}
        \vspace{-1ex}
        \caption{\textcolor{Fix2}{Comparison on the hyper-parameter setting for language-pair agnostic scenarios.} Average\textcolor{Fix2}{d F1 score} with \textsc{InfoXLM} and \textsc{XLM-R} trained on synthetic data for different hyper-parameter}
        \label{fig.alpha}

\end{figure}

Pulling away the representation between single sentence inputs and concatenated inputs can reduce the influence of monolingual features brought by PLM and can improve the overall performance.
While there is a trade-off between reducing the effect of monolingual features and destroying the representative ability of the encoder which depends on the parameter $\alpha$.
To understand the impact of $\alpha$, we train \textsc{InfoXLM} and \textsc{XLM-R} based models with different $\alpha$ and evaluate their performance.
As shown in Figure \ref{fig.alpha},  with the increase of $\alpha$, the average performance rise first and then falls, and the best $\alpha$ selection is $0.05$ and $0.1$ for \textsc{InfoXLM} and \textsc{XLM-R}, respectively.
The best $\alpha$ selected for \textsc{XLM-R} based model is quite large than \textsc{InfoXLM} based model.
The reason lies in comparison to \textsc{InfoXLM}, XLM-R only trained by MLM during the pretraining period which means they receive more impact on monolingual features.
The result also verifies our assumption that $\alpha$ should be kept on a small scale to avoid destroying the representative ability of models.
\subsection{Language-pair Specific}
\begin{figure}
    \centering
        \begin{tikzpicture}
            \pgfplotsset{set layers}
            \pgfplotsset{every x tick label/.append style={font=\small}}
            \pgfplotsset{every y tick label/.append style={font=\small}}
            \begin{axis}[
                height=0.65 * \columnwidth,
                width=\columnwidth,
                xmin=0, xmax=0.4,
                ymin=18, ymax=26,
                xtick={0, 0.05, 0.1, 0.2, 0.4},
                xticklabels={$0$, $0.05$, $0.1$, $0.2$, $0.5$},
                ytick={18,22,26},
                yticklabels={$18$, $22$, $26$},
                ymajorgrids=true,
                grid style=dashed,
                legend cell align=left,
                legend style={
                    at={(0.85, 1.05)},
                    anchor=south east,
                    font=\tiny,
            		legend columns=2},
            	every axis plot/.append style={thick},
                ]
            \addplot[
                color=blue,
                mark=*
                ]
                plot coordinates {
                    (0, 20.8)
                    (0.01, 20.9)
                    (0.05, 24.9)
                    (0.1, 23.3)
                    (0.2, 21.5)
                    (0.4, 22.8)
                    
            };
            \addlegendentry{\textsc{InfoXLM-ZhEn}}
            
            \addplot[
                color=blue!80!white,
                mark=o
                ]
                plot coordinates {
                    (0, 20.4)
                    (0.01, 22.7)
                    (0.05, 21.5)
                    (0.1, 21.2)
                    (0.2, 20.7)
                    (0.4, 21.6)
            };
            \addlegendentry{\textsc{InfoXLM-EnDe}}
            
            \addplot[
                color=green!80!black,
                mark=*
                ]
                plot coordinates {
                    (0, 19.1)
                    (0.01, 20.7)
                    (0.05, 22.7)
                    (0.1, 23.6)
                    (0.2, 23.9)
                    (0.4, 22.4)
                    
                };
            \addlegendentry{\textsc{XLM-R-ZhEn}}
                        \addplot[
                color=green!80!black,
                mark=o
                ]
                plot coordinates {
                    (0, 19.3)
                    (0.05, 21.8)
                    (0.1, 21.9)
                    (0.2, 20.8)
                    (0.4, 21.1)
                    
                };
            \addlegendentry{\textsc{XLM-R-EnDe}}
            \end{axis}
        \end{tikzpicture}
        \vspace{-1ex}
        \caption{\textcolor{Fix2}{Comparison on the hyper-parameter setting for language-pair specific scenarios.} Average\textcolor{Fix2}{d F1 score} with \textsc{InfoXLM} and \textsc{XLM-R} trained on synthetic data for different hyper-parameter\textcolor{Fix2}{s.}}
        \label{fig.lp_alpha}

\end{figure}
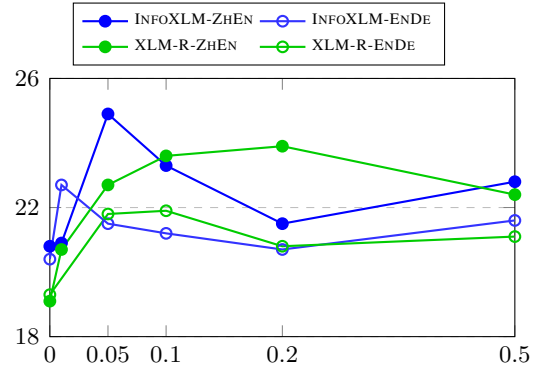
We are further curious about whether different language pairs may have different $\alpha$ settings for the best performance.
We first split the synthetic data into En-De and Zh-En subsets.
Then, we train \textsc{InfoXLM} and \textsc{XLM-R} based models with different $\alpha$ values, and evaluate them on MQM datasets.
As shown in Figure \ref{fig.lp_alpha}, the best $\alpha$ selection for En-De is always quite small than it for Zh-En when using the same backbones (InfoXLM-En-De: 0.01, InfoXLM-Zh-En: 0.05, XLM-R-En-De: 0.1, XLM-R-Zh-En 0.2). We attribute this phenomenon to the fact that En and De are similar languages, thus the language model is easier to align the semantics of En and De than those of En and Zh.

\section{Size of Synthetic Data}
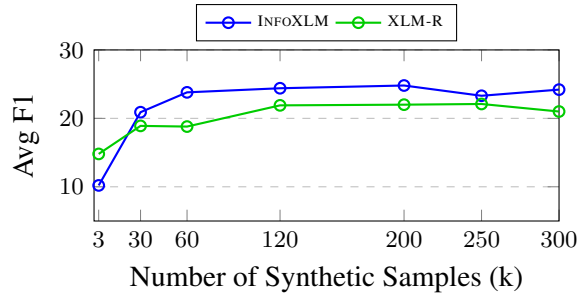
\begin{figure}[t]
    \centering
    \scalebox{1.0}{
        \begin{tikzpicture}
            \pgfplotsset{set layers}
            \pgfplotsset{every x tick label/.append style={font=\small}}
            \pgfplotsset{every y tick label/.append style={font=\small}}
            \begin{axis}[
                height=0.5 * \columnwidth,
                width=\columnwidth,
                xmin=0, xmax=300,
                ymin=5, ymax=30,
                xtick={3,  30, 60, 120, 200, 250, 300},
                xticklabels={$3$,  $30$, $60$, $120$, $200$, $250$, $300$},
                ytick={10,20,30},
                yticklabels={$10$,$20$, $30$},
                ylabel={Avg F1},
                xlabel={Number of \textcolor{Fix2}{Synthetic} Samples (k)},
                ymajorgrids=true,
                grid style=dashed,
                legend cell align=left,
                legend style={
                    at={(0.5, 1.05)},
                    anchor=south,
                    font=\tiny,
            		legend columns=3},
            	every axis plot/.append style={thick},
                ]
            \addplot[
                color=blue,
                mark=o
                ]
                plot coordinates {
                    (3, 10.2)
                    (30, 20.9)
                    (60, 23.8)
                    (120, 24.4)
                    (200, 24.8)
                    (250, 23.3)
                    (300, 24.2)
                
            };
            \addlegendentry{\textsc{InfoXLM}}
            
            \addplot[
                color=green!80!black,
                mark=o
                ]
                plot coordinates {
                    (3, 14.8)
                    (30, 18.9)
                    (60, 18.8)
                    (120, 21.9)
                    (200, 22.0)
                    (250, 22.1)
                    (300, 21.0)
                };
            \addlegendentry{\textsc{XLM-R}}
                
            \end{axis}
        \end{tikzpicture}
        }
        \vspace{-1ex}
        \caption{\textcolor{Fix2}{Comparison on the number of synthetic samples during training. We use different numbers of synthetic examples (x-axis), and collect the averaged F1 score (y-axis).}}
        \label{figure.syn_size}

\end{figure}
Figure \ref{figure.syn_size} shows the influence of synthetic data for training our models. 
As we can see, there is no significant performance improvement when the number of samples is over 60k and 120k for \textsc{InfoXLM} and \textsc{XLM-R}, respectively.
Compared with using a few training samples (3k), we can observe that increasing the amount of synthetic data brings an improvement of 13.8 and 7.1 F1 scores on average for \textsc{InfoXLM} and \textsc{XLM-R}.
This demonstrates the power of our methods in generating pseudo data and \textsc{InfoXLM} benefits more from synthetic data than \textsc{XLM-R}. 
Currently, researchers often use synthetic data for training evaluation models~\cite{zhong2022towards}.
However, the distribution gap between pseudo data and real samples troubles the training of the evaluation model, which easily leads to model overfitting. 
We believe that, in the future, generating in-distribution pseudo examples is quite a worth-thinking direction.
\section{Model Predictions}
%

\begin{CJK*}{UTF8}{gbsn}
To show the power of our models in real scenarios, we recruit annotators to identify whether the model's predictions are good or not.
Table \ref{table_prediction} shows the predictions of our models and those of~\newcite{vamvas-sennrich-2022-little} in some cases. 
Also, we list some significant bad cases in table \ref{table_badcase}, which should be paid more attention to in the future.
Specifically, compared with existing methods~\cite{vamvas-sennrich-2022-little}, our models can identify 
a span of complete addition/omission errors. 
Besides, our models have the ability to find more granular errors (\textit{e.g.},     ``country'' vs. ``大国''). 
Nevertheless, our models still have major flaws. The model has serious problems with semantic understanding, especially terminologies or specific meanings (\textit{e.g.}, ``long tunnel short strike'' vs. ``长隧短打''). 
It cannot accurately judge the meaning of such words. 
Coincidentally, we also find that  mistranslation errors are mainly raised in translating terminologies. 
We believe that introducing external knowledge databases~\citep[\textit{e.g.}, multilingual knowledge graph,][]{yao2019kg} to our FG-TED models is expected to help identify the mistranslation errors~\cite{amrhein2022aces}, and we would like to leave this open problem to future work.

\end{CJK*}

\begin{CJK*}{UTF8}{gbsn}
    \begin{table*}[t]
        \centering
        \scalebox{0.9}{
            \begin{tabularx}{\textwidth}{lX}
                \toprule
                
                \textbf{Source:} & 由此可以看到，大兴国际机场的通航，不只是增加了几条跑道和几架飞机而已，无论是从企业竞争层面、消费者体验层面，还是从区域航空一体化、京津冀发展一体化层面，都具有深远影响和积极意义，不仅事关北京航运产业的长远布局和发展，为京津冀广大居民的出行提供多样化选择，推动北京航运业服务质量进一步提升，更有利于促进京津冀一体化发展进程，拉动大兴机场周边地区的经贸发展。\\
                \cdashline{1-2}[1pt/2.5pt]\noalign{\vskip 0.5ex}
                \textbf{Target:} & From this, it can be seen that the opening of Daxing International Airport has not only added a few runways and a few aircraft. Whether from the level of enterprise competition from the level of enterprise competition, the level of consumer experience, or from the level of regional aviation integration and the development and integration of Beijing, Tianjin and Hebei, it has far-reaching influence and positive significance, which is not only related to the long-term layout and development of Beijing's shipping industry. \\
                \cdashline{1-2}[1pt/2.5pt]\noalign{\vskip 0.5ex}
                \textbf{MQM Label:} & From this, it can be seen that the opening of Daxing International Airport has not only added a few runways and a few aircraft. Whether from the level of enterprise competition from the level of enterprise competition, the level of consumer experience, or from the level of regional aviation integration and the development and integration of Beijing, Tianjin and Hebei, it has far-reaching influence and positive significance, which is not only related to the long-term layout and development of Beijing's shipping industry. \textcolor{cyan}{It also provides residents in the Beijing-Tianjin-Hebei region more varied means of transportation and promotes the improvement of Beijing's shipping industry. It will benefit the integrated development of Beijing-Tianjin-Hebei and the economic and trade development of the airport's surrounding areas.} \\
                \cdashline{1-2}[1pt/2.5pt]\noalign{\vskip 0.5ex}
                \textbf{Our Label:} & 由此可以看到，大兴国际机场的通航，不只是增加了几条跑道和几架飞机而已，无论是从企业竞争层面、消费者体验层面，还是从区域航空一体化、京津冀发展一体化层面，都具有深远影响和积极意义，不仅事关北京航运产业的长远布局和发展，\textcolor{cyan}{为京津冀广大居民的出行提供多样化选择，推动北京航运业服务质量进一步提升，更有利于促进京津冀一体化发展进程，拉动大兴机场周边地区的经贸发展。} \\
                
                \midrule
                
                \textbf{Source:} & 大兴机场的配套交通设施非常完善，除了北京市区直达大兴机场的新机场线外，还有多条高速公路联通大兴机场与周边地区。\\
                \cdashline{1-2}[1pt/2.5pt]\noalign{\vskip 0.5ex}
                \textbf{Target:} &  The supporting transportation facilities of Daxing Airport are very perfect. In addition to the new airport line directly to Daxing Airport in Beijing, there are also a number of expressways connecting Daxing Airport with the surrounding areas. \\
                \cdashline{1-2}[1pt/2.5pt]\noalign{\vskip 0.5ex}
                \textbf{MQM Label:} & The supporting transportation facilities of Daxing Airport are very perfect. In addition to the new airport line directly to Daxing Airport in\textcolor{cyan}{in downtown} Beijing, there are also a number of expressways connecting Daxing Airport with the surrounding areas. \\
                \cdashline{1-2}[1pt/2.5pt]\noalign{\vskip 0.5ex}
                \textbf{Our Label:} & 大兴机场的配套交通设施非常完善，除了北京\textcolor{cyan}{市区}直达大兴机场的新机场线外，还有多条高速公路联通大兴机场与周边地区。\\
                
                \midrule
                
                \textbf{Source:} & 中国海军在亚丁湾、索马里海域护航已逾10年，累计完成1200余批6700余艘船舶护航任务。\\
                \cdashline{1-2}[1pt/2.5pt]\noalign{\vskip 0.5ex}
                \textbf{Target:} & The Chinese Navy has been escorting the waters of the Gulf of Aden and Somalia for more than 10 years, and has completed more than 1,200 batches of more than 6,700 ships. \\
                \cdashline{1-2}[1pt/2.5pt]\noalign{\vskip 0.5ex}
                \textbf{MQM Label:} & The Chinese Navy has been \textcolor{cyan}{escorting ships} the waters of the Gulf of Aden and Somalia for more than 10 years, and has completed more than 1,200 batches of more than 6,700 ships.\\
                \cdashline{1-2}[1pt/2.5pt]\noalign{\vskip 0.5ex}
                \textbf{Our Label:} & 中国海军在亚丁湾、索马里海域护\textcolor{cyan}{航}已逾10年，累计完成1200余批6700余艘船舶护航任务。\\
                
                \bottomrule
                 
            \end{tabularx}
        }

        \caption{Some examples of the MQM labels and our labels for omission errors.}
        \label{table_mqm}
    \end{table*}
\end{CJK*}

\begin{CJK*}{UTF8}{gbsn}
    \begin{table*}[t]
        \centering
        \scalebox{0.9}
        {
            \begin{tabularx}{\textwidth}{llX}
                \toprule
                \multirow{2}{*}{Ours} & \textbf{Source:} & \textcolor{red}{中新网7月26日电据外媒报道,}7月18日, 法国南特\textcolor{red}{的宗教历史建筑瑰宝}、著名的圣彼得与圣保罗大教堂, 在一场火灾当中受到严重损毁, 大管风琴完全毁坏。 \\
                & \textbf{Target:} & The cathedral of Saint Peter and Saint Paul in Nantes, France, was severely damaged in a fire on July 18, with the large organ completely destroyed. \\
                \cdashline{1-3}[1pt/2.5pt]\noalign{\vskip 0.5ex}
                \multirow{2}{*}{\textsc{Contrastive Conditioning}} & \textbf{Source:} & \textcolor{red}{中新网7月26日电据外媒}报道，7月18日, 法国南特\textcolor{red}{的宗教历史建筑}瑰宝、\textcolor{red}{著名的}圣彼得与圣保罗大教堂, 在一场火灾当中受到严重损毁, 大管风琴完全毁坏。 \\
                & \textbf{Target:} & The cathedral of Saint \textcolor{red}{Peter} and Saint Paul in Nantes, France, was severely damaged in a fire on July \textcolor{red}{18}, with the large organ completely destroyed. \\
                \cdashline{1-3}[1pt/2.5pt]\noalign{\vskip 0.5ex}
                \multirow{2}{*}{Ground-Truth} & \textbf{Source:} & \textcolor{red}{中新网7月26日电据外媒报道,}7月18日, 法国南特\textcolor{red}{的宗教历史建筑瑰宝}、\textcolor{red}{著名的}圣彼得与圣保罗大教堂, 在一场火灾当中受到严重损毁, 大管风琴完全毁坏。 \\
                & \textbf{Target:} & The cathedral of Saint Peter and Saint Paul in Nantes, France, was severely damaged in a fire on July 18, with the large organ completely destroyed. \\
                \midrule
                \multirow{2}{*}{Ours} & \textbf{Source:} & \textcolor{red}{中新网7月26日电综合俄罗斯卫星网报道}， 7月26日是俄罗斯海军日，庆祝海军的主要阅兵式在圣彼得堡和喀琅施塔得举行。 \\
                & \textbf{Target:} & Russia's Navy Day is celebrated on July 26, with military parades in Saint Petersburg and Kronstadt. \\
                \cdashline{1-3}[1pt/2.5pt]\noalign{\vskip 0.5ex}
                \multirow{2}{*}{\textsc{Contrastive Conditioning}} & \textbf{Source:} & 中新网\textcolor{red}{7月26日}电\textcolor{red}{综合}俄罗斯卫星网报道， 7月26日是\textcolor{red}{俄罗斯海军日}，庆祝海军的\textcolor{red}{主要}阅兵式在圣彼得堡和喀琅施塔得\textcolor{red}{举行}。 \\
                & \textbf{Target:} & \textcolor{red}{Russia'}s Navy Day is celebrated on July 26, with \textcolor{red}{military} parades in Saint Petersburg and Kronstadt. \\
                \cdashline{1-3}[1pt/2.5pt]\noalign{\vskip 0.5ex}
                \multirow{2}{*}{Ground-Truth} & \textbf{Source:} & \textcolor{red}{中新网7月26日电综合俄罗斯卫星网报道}， 7月26日是俄罗斯海军日，庆祝海军的主要阅兵式在圣彼得堡和喀琅施塔得举行。 \\
                & \textbf{Target:} & Russia's Navy Day is celebrated on July 26, with military parades in Saint Petersburg and Kronstadt. \\
                \midrule
                \multirow{2}{*}{Ours} & \textbf{Source:} &  中国已经成为创新和知识产权\textcolor{red}{大}国。\\
                & \textbf{Target:} & China has become a country of innovation and intellectual property rights. \\
                \cdashline{1-3}[1pt/2.5pt]\noalign{\vskip 0.5ex}
                \multirow{2}{*}{\textsc{Contrastive Conditioning}} & \textbf{Source:} &  中国已经成为创新和知识产权大国。\\
                & \textbf{Target:} & China has become a country of innovation and intellectual property rights. \\
                \cdashline{1-3}[1pt/2.5pt]\noalign{\vskip 0.5ex}
                \multirow{2}{*}{Ground-Truth} & \textbf{Source:} &  中国已经成为创新和知识产权\textcolor{red}{大}国。\\
                & \textbf{Target:} & China has become a country of innovation and intellectual property rights. \\
                \midrule
                \multirow{2}{*}{Ours} & \textbf{Source:} & 这项研究于24日在《美国医学会杂志》上刊登。\\
                & \textbf{Target:} & The research was published on \textcolor{red}{September} 24th in the Journal of the American Medical Association. \\
                \cdashline{1-3}[1pt/2.5pt]\noalign{\vskip 0.5ex}
                \multirow{2}{*}{\textsc{Contrastive Conditioning}} & \textbf{Source:} & 这项研究于24日在《美国医学会杂志》上刊登。\\
                & \textbf{Target:} & The research was published on September 24th in the Journal of the American Medical Association. \\
                \cdashline{1-3}[1pt/2.5pt]\noalign{\vskip 0.5ex}
                \multirow{2}{*}{Ground-Truth} & \textbf{Source:} & 这项研究于24日在《美国医学会杂志》上刊登。\\
                & \textbf{Target:} & The research was published on \textcolor{red}{September} 24th in the Journal of the American Medical Association. \\
                
                \bottomrule
            \end{tabularx}
        }
        \caption{Case study for our model and \textsc{Contrastive Conditioning}~\cite{vamvas-sennrich-2022-little}.}
        \label{table_prediction}
    \end{table*}
\end{CJK*}

\begin{CJK*}{UTF8}{gbsn}
    \begin{table*}[t]
        \centering
        \scalebox{0.9}{
            \begin{tabularx}{\textwidth}{lX}
                \toprule
                \textbf{Source} & 竖井则是高黎贡山隧道最重要的辅助坑道,肩负增加作业面实现“长隧短打”和后期铁路运营通风的重要任务。 \\
                \textbf{Target} & The shaft is the most important auxiliary tunnel of Gaoligong Mountain Tunnel, shouldering the important task of increasing the operation surface to realize \textcolor{red}{"long tunnel short strike}" and later railway operation ventilation. \\
                \midrule
                \textbf{Source} & 要坚持标本兼\textcolor{red}{治,完善长效机制},织密扎牢制度笼子,强化制度刚性约束,确保各项工作有章可循、有规可依,真正用制度管人\textcolor{red}{管事管权。} \\
                \textbf{Target} & We should adhere to both symptoms and treatment, improve the long-term mechanism, tighten the system cage, strengthen the rigid constraints of the system, ensure that all work is governed by rules and regulations, and truly use the system to manage people and manage power. \\
                \midrule 
                \textbf{Source} & 走过70年, 中国正青春、昂扬、\textcolor{red}{风华正茂,无论}国际风云如何变化,我们都将坚持自己的初心。\\
                \textbf{Target} & After 70 years, China is young, exuberant and prosperous. No matter how the international situation changes, we will adhere to our first heart. \\
                \bottomrule                 
            \end{tabularx}

        }

        \caption{Case study for our model.}
        \label{table_badcase}
    \end{table*}
\end{CJK*}

\begin{table*}[t]
    \centering
    \small
    \scalebox{0.7}{

        \begin{tabular}{lccccccccccccc}
            \toprule
            \multirow{3}{*}{\textbf{Model}} & \multicolumn{6}{c}{\textbf{En-De}} & \multicolumn{6}{c}{\textbf{En-Zh}} & \multirow{3}{*}{\textbf{Avg MCC}}\\  
           \cmidrule(l{3pt}r{3pt}){2-7}
           \cmidrule(l{3pt}r{3pt}){8-13}
            & \multicolumn{3}{c}{\textbf{Source}} & \multicolumn{3}{c}{\textbf{Target}} & \multicolumn{3}{c}{\textbf{Source}} & \multicolumn{3}{c}{\textbf{Target}} & \\
           \cmidrule(l{3pt}r{3pt}){2-4}
           \cmidrule(l{3pt}r{3pt}){5-7}
           \cmidrule(l{3pt}r{3pt}){8-10}
           \cmidrule(l{3pt}r{3pt}){11-13}
            
            & \textbf{F1\_OK} & \textbf{F1\_BAD} & \textbf{MCC} & \textbf{F1\_OK} & \textbf{F1\_BAD} & \textbf{MCC} & \textbf{F1\_OK} & \textbf{F1\_BAD} & \textbf{MCC} & \textbf{F1\_OK} & \textbf{F1\_BAD} & \textbf{MCC} & \\
        \midrule
        \textsc{Baseline} & 92.4 & 39.3 & 32.3 & 91.1 & 45.5 & 37.0 & 75.1 & 39.4 & 24.1  & 72.3 & 42.6 & 24.7 & 29.5 \\
        \midrule
        \multicolumn{14}{c}{\textcolor{Fix2}{\textit{Backbone}: \textsc{XLM-R}}}  \\
        \cdashline{1-14}[1pt/2.5pt]\noalign{\vskip 0.5ex}         
        
        \textsc{+ FT} & 89.6 & 35.2 & 25.5 & 91.7 & 47.1 & 38.9 & 75.0 & 49.2 & 27.8 & 75.4 & 55.3 & 33.3 & 31.4 \\
        \textsc{+ FT*} &89.7&35.2&25.4&91.5&46.6&38.2&73.8&49.7&28.2&73.9&56.4&34.5&31.6 \\
        \textsc{+ SYN + FT} & 89.9&36.3&26.7&91.9&49.3&41.3&77.0&50.1&29.5&78.3&56.9&36.5&33.5 \\
        \textsc{+ SYN + FT*} &89.7&38.2&28.8&91.7&50.1&41.9&76.5&50.3&29.6&77.7&57.0&36.4&34.2  \\
        \textsc{+ SYN* + FT} & 90.0 & 35.2 & 25.6 & 92.1 & 49.6 & 41.8 & 78.2 & 50.1 & 30.0  & 78.6 & 56.1 & 35.6 & 33.3 \\
        \textsc{+ SYN* + FT*} & 90.2 & 37.8 & 28.5 & 91.7 & 50.2 & 41.9 & 77.3 & 50.5 & 30.1 & 77.9 & 56.7 & 36.1 & 34.2 \\
        
        \midrule
        \multicolumn{14}{c}{\textcolor{Fix2}{\textit{Backbone}: \textsc{InfoXLM}}}  \\
        \cdashline{1-14}[1pt/2.5pt]\noalign{\vskip 0.5ex}        
        \textsc{+ FT} & 93.1 & 39.8 & 33.6 & 92.3 & 46.5 & 39.6 & 79.4 & 50.7 & 31.3 & 78.9 & 56.4 & 36.1 & 35.2 \\
        \textsc{+ FT*}&93.3&38.9&33.4&92.6&46.1&40.1&79.9&50.6&31.4&79.4&56.4&36.4&35.3 \\
        \textsc{+ SYN + FT} & 93.1&40.1&33.9&92.5&46.4&39.9&78.6&50.2&30.3&78.7&55.6&35.1&34.8 \\
        \textsc{+ SYN + FT*}&93.2&42.8&36.5&92.6&47.8&41.4&78.1&50.7&30.7&78.2&55.6&34.8&35.8 \\
        \textsc{+ SYN* + FT} & 93.1 & 42.0 & 35.4 & 92.6 & 49.3 & 42.6 & 78.9 & 50.6 & 30.9 & 48.7 & 55.8 & 35.4 & 36.1 \\ 
        \textsc{+ SYN* + FT*} & 93.3 & 40.4 & 34.5 & 92.7 & 47.8 & 41.7 & 79.4 & 50.2 & 30.7 & 79.5 & 55.7 & 35.7 & 35.7 \\
        \bottomrule
        \end{tabular}
    }
        \caption{The detailed results of our methods on the word-level QE dataset for En-De and En-Zh directions.}

    \label{table_word_level_detailed}
\end{table*}

\end{document}